\documentclass[journal]{IEEEtran}

\usepackage{times}  
\usepackage{helvet} 
\usepackage{courier}  
\usepackage[hyphens]{url}  
\usepackage{graphicx} 
\usepackage{caption} 
\usepackage{algorithm}
\usepackage[noend]{algpseudocode}

\usepackage{url}            
\usepackage{booktabs}       

\usepackage[switch]{lineno}

\usepackage{amssymb,amsmath,amsthm,bm}

\newtheorem{theorem}{Theorem}
\newtheorem{definition}{Definition}
\newtheorem{lemma}{Lemma}

\usepackage{comment}
\usepackage{enumitem}

\usepackage{xcolor}

\def\vn{{\bm{n}}}

\def\vu{{\bm{u}}}

\def\vw{{\bm{w}}}
\def\vx{{\bm{x}}}
\def\vy{{\bm{y}}}

\def\mA{{\bm{A}}}
\def\mB{{\bm{B}}}

\def\mI{{\bm{I}}}

\def\mU{{\bm{U}}}

\def\mW{{\bm{W}}}

\def \cM {{\mathcal{M}}}
\def \cS {{\mathcal{S}}}
\def \cR {{\mathcal{R}}}
\def \PP {{\mathbb{P}}}
\def \RR {{\mathbb{R}}}
\def \EE {{\mathbb{E}}}

\def \bX {{\bm X}}

\def \Cb {{\mathbf{C}}}

\usepackage{verbatim}

\begin{document}

\title{Deep Learning with Data Privacy via Residual Perturbation}

\author{Wenqi~Tao*, Huaming~Ling*, Zuoqiang~Shi, 
        and~Bao~Wang
\thanks{B. Wang is with the Department
of Mathematics and Scientific Computing and Imaging Institute, University of Utah, Salt Lake City,
UT, 84112 USA. E-mail: wangbaonj@gmail.com.}
\thanks{W. Tao, H. Ling, and Z. Shi are with the Department of Mathematics, Tsinghua University, Beijing 100084. E-mail: twq17@mails.tsinghua.edu.cn, linghm18@mails.tsinghua.edu.cn, zqshi@mail.tsinghua.edu.cn. Wenqi Tao and Huaming Ling contribute equally.}
}

\markboth{Journal of \LaTeX\ Class Files,~Vol.~14, No.~8, August~2015}%
{Shell \MakeLowercase{\textit{et al.}}: Bare Demo of IEEEtran.cls for IEEE Journals}

\maketitle

\begin{abstract}
Protecting data privacy in deep learning (DL) is of crucial importance. Several celebrated privacy notions have been established and used for privacy-preserving DL. However, many existing mechanisms achieve privacy at the cost of significant utility degradation and computational overhead. In this paper, we propose a stochastic differential equation-based \emph{residual perturbation} for privacy-preserving DL, which injects Gaussian noise into each residual mapping of ResNets. Theoretically, we prove that residual perturbation guarantees differential privacy (DP) and reduces the generalization gap of DL. Empirically, we show that residual perturbation is computationally efficient and outperforms the state-of-the-art differentially private stochastic gradient descent (DPSGD) in utility maintenance {without sacrificing membership privacy}. 
\end{abstract}


\IEEEpeerreviewmaketitle

\vspace{-0.4cm}
\section{Introduction}
\label{sec:introduction}
Many high-capacity deep neural networks (DNNs) are trained with private data, including medical images and financial transaction data \cite{Yuen:2011,Feng:2017,Liu:2017}. DNNs usually overfit and can memorize the private training data, making training DNNs exposed to privacy leakage~\cite{Fredrikson:2015,shokri2017membership,salem2018ml,yeom2018privacy,sablayrolles2018d}. Given a pre-trained DNN, the membership inference attack can determine if an instance is in the training set based on DNN's response \cite{fredrikson2014privacy,shokri2017membership,salem2018ml}; the model extraction attack can learn a surrogate model that matches the target model, given the adversary only black-box access to the target model~\cite{tramer2016stealing,gong2018attribute}; the model inversion attack can infer certain features of a given input from the output of a target model \cite{fredrikson2015model,al2016reconstruction}; the attribute inference attack can deanonymize the anonymized data \cite{gong2016you,zheng2018data}.

Machine learning (ML) with data privacy is crucial in many applications~\cite{lindell2000privacy,barreno2006can,hesamifard2018privacy,bae2019anomigan}. Several algorithms have been developed to reduce privacy leakage, including differential privacy (DP) \cite{dwork2006calibrating} and federated learning (FL) \cite{mcmahan2016communication,konevcny2016federated}, and $k$-anonymity \cite{sweeney2002k,el2008protecting}. Objective, output, and gradient perturbations are popular approaches for ML with DP guarantee  \cite{chaudhuri2011differentially,bassily2014private,Shokri:2015,abadi2016deep,bagdasaryan2019differential}. FL trains centralized ML models, through gradient exchange, with the training data being distributed on the edge devices. However, the gradient exchange can still leak privacy~\cite{zhu2019deep,wang2019beyond}. Most of the existing privacy is achieved at a tremendous sacrifice of utility. Moreover, training ML models using the state-of-the-art DP stochastic gradient descent (DPSGD) leads to tremendous computational cost due to the requirement of computing and clipping the per-sample gradient \cite{Abadi:2016}. \emph{It remains a great interest to develop new privacy-preserving ML algorithms without the excessive computational overhead or degrade the ML models' utility.}

\vspace{-0.4cm}
\subsection{Our Contribution}\vspace{-0.1cm}
We propose \emph{residual perturbation} for privacy-preserving deep learning (DL). At the core of residual perturbation is injecting Gaussian noise to each residual mapping \cite{he2016deep}, and the residual perturbation is theoretically motivated by the stochastic differential equation (SDE) theory. The major advantages of residual perturbation are summarized below:
\begin{itemize}[leftmargin=*]
\item It guarantees DP and can protect the membership privacy of the training data almost perfectly without sacrificing ResNets' utility, even improving the classification accuracy. 

\item It has fewer hyperparameters to tune than DPSGD. Moreover, it is computationally much more efficient than DPSGD; the latter is computationally prohibitive due to the requirement of computing the per-sample gradient.
\end{itemize}

\vspace{-0.4cm}
\subsection{Related Work}\vspace{-0.1cm}
Improving the utility of ML models with DP guarantees is an important task. PATE \cite{Papernot:2017,Papernot:2018} uses semi-supervised learning {accompanied by} model transfer between the ``student'' and ``teacher'' models to enhance utility. Several variants of the DP notions have also been proposed to improve the {theoretical} privacy budget and sometimes can also improve the resulting model's utility
~\cite{abadi2016deep,mironov2017renyi,wang2018subsampled,dong2019gaussian}. Some post-processing techniques have also been developed to improve the utility of ML models with negligible computational overhead \cite{wang2019dp,liang2020exp}.

{The tradeoff between privacy and utility of ML models has been studied in several different contexts. In \cite{9799176}, the authors identify the factors that impact the optimal tradeoff of accuracy and privacy of differentially private DL models by empirically evaluating the commonly used privacy-preserving ML libraries, including PyTorch Opacus and Tensorflow privacy. The authors of \cite{jagielski2020auditing} investigate whether DPSGD offers better privacy in practice than what is guaranteed by its state-of-the-art theoretical privacy analysis --- by taking a quantitative, empirical approach to auditing the privacy afforded by specific implementations of differentially private algorithms. In particular, they estimate an empirical DP lower bound under the data poisoning and membership inference attacks. They further compare this empirical lower bound with the DP budget obtained by the state-of-the-art analytic tools \cite{abadi2016deep,dong2019gaussian,mironov2019r,yu2019differentially}. The approach presented in \cite{jagielski2020auditing} can demonstrate if a given algorithm with a given DP budget is sufficiently private and provides empirical evidence to guide the development of theoretical DP analysis. The privacy-utility tradeoff for different privacy measures in both global and local privacy has been studied in \cite{zhong2022privacy}, aiming to determine the minimal privacy loss at a fixed utility expense. The paper \cite{mivule2012towards} investigates a DP ensemble classifier approach that seeks to preserve data privacy while maintaining an acceptable level of utility. For the linear regression model, the privacy-utility tradeoff has been studied by enforcing the privacy of the input data via random projection and additive noise \cite{showkatbakhsh2018privacy}.}

Gaussian noise injection in residual learning has been used to improve the robustness of ResNets \cite{rakin2018parametric,wang2019resnets,liu2019neural}. In this paper, we inject Gaussian noise into each residual mapping to achieve data privacy {with even improved generalization} instead of adversarial robustness. {Several related studies focus on disentangling adversarial robustness and generalization in the literature. Some initial hypotheses state that there exists an inherent tradeoff between robustness and accuracy, i.e., an ML model cannot be both more robust and more accurate than others \cite{su2018robustness,tsipras2018robustness} based on the comparisons of different models with different architectures or training strategies and theoretical arguments on a toy dataset, respectively. In \cite{stutz2019disentangling}, the authors disentangle adversarial robustness and generalization, leveraging the low-dimensional manifold assumption of the data. In particular, through a detailed analysis of the regular and on-manifold \cite{gilmer2018adversarial,zhao2018generating} adversarial examples of the same type as the given class --- the adversarial examples that leave and stay in the manifold of the given class of data, the authors of \cite{stutz2019disentangling} conclude that both robust and accurate ML models are possible. In particular, on-manifold robustness is essentially the generalization of the ML model, which can be improved via on-manifold adversarial training. In \cite{pmlr-v119-raghunathan20a}, the authors precisely characterize the effects of data augmentation on the standard error (corresponding to the natural accuracy of the model) in linear regression when the optimal linear predictor has zero standard and robust error (corresponding to the model's robust accuracy). Moreover, it is shown in \cite{pmlr-v119-raghunathan20a} that robust self-training \cite{carmon2019unlabeled,najafi2019robustness,alayrac2019labels} can improve the linear regression model's robustness without sacrificing its standard error by leveraging extra unlabelled data. Our approach of the ensemble of noise-injected neural networks provides another feasible avenue to develop ML models that are both robust and accurate.}




\vspace{-0.4cm}
\subsection{Organization}\vspace{-0.1cm}
We organize this paper as follows: In Sec.~\ref{sec:Algorithm}, we introduce the residual perturbation for privacy-preserving DL. In Sec.~\ref{sec:Theory}, we present the generalization and DP guarantees for residual perturbation. In Sec.~\ref{sec:Experiments}, we numerically verify the efficiency of the residual perturbation in protecting data privacy without degrading the underlying models' utility. Technical proofs and some more experimental details and results are provided in Secs.~\ref{sec:proofs} and \ref{sec:appendix:strategyII}. We end up with concluding remarks.

\vspace{-0.4cm}
\subsection{Notations}\vspace{-0.1cm}
We denote scalars by lower or upper case letters; vectors and matrices by lower and upper case bold face letters, respectively. For a vector $\vx=(x_1,\ldots,x_d)\in \mathbb{R}^d$, we use $\|\vx\|_2 = {(\sum_{i=1}^d |x_i|^2)^{1/2}}$ to denote its $\ell_2$ norm. For a matrix $\mA$, we use $\|\mA\|_{2}$ to denote its induced norm by the vector $\ell_{2}$ norm. We denote the standard Gaussian in $\RR^d$ as $\mathcal{N}(\mathbf{0}, \mI)$ with $\mI\in \RR^{d\times d}$ being the identity matrix. The set of (positive) real numbers is denoted as ($\RR^+$) $\RR$. We use $B(\mathbf{0}, R)$ to denote the ball centered at $\mathbf{0}$ with radius $R$.

\vspace{-0.2cm}
\section{Algorithms}\label{sec:Algorithm}
\vspace{-0.1cm}
\subsection{Deep Residual Learning and Its Continuous Analogue}
\begin{figure*}[t!]
\centering
\begin{tabular}{ccccc}
\hskip -0.2cm\includegraphics[clip, trim=0cm 1cm 0cm 0cm,width=0.33\columnwidth]{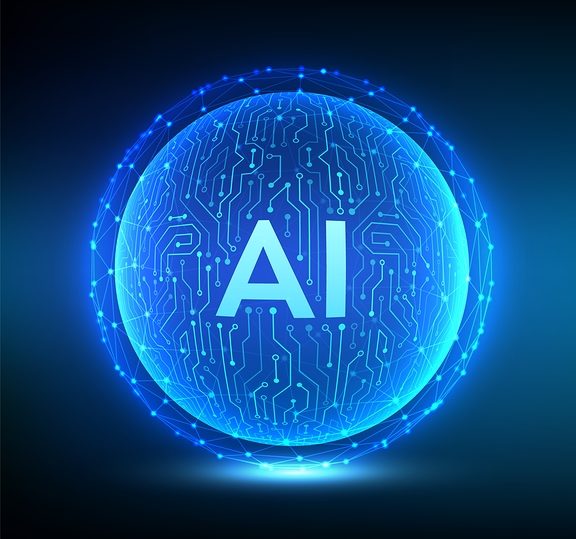}&
\hskip -0.2cm\includegraphics[clip, trim=0cm 1cm 0cm 0cm,width=0.33\columnwidth]{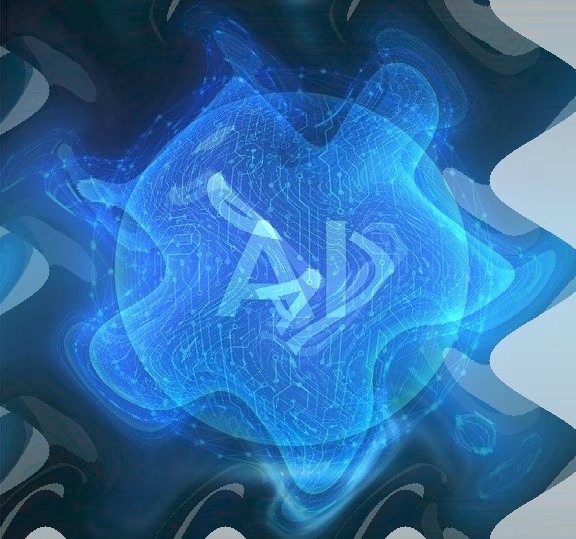}&
\hskip -0.2cm\includegraphics[clip, trim=0cm 1cm 0cm 0cm,width=0.33\columnwidth]{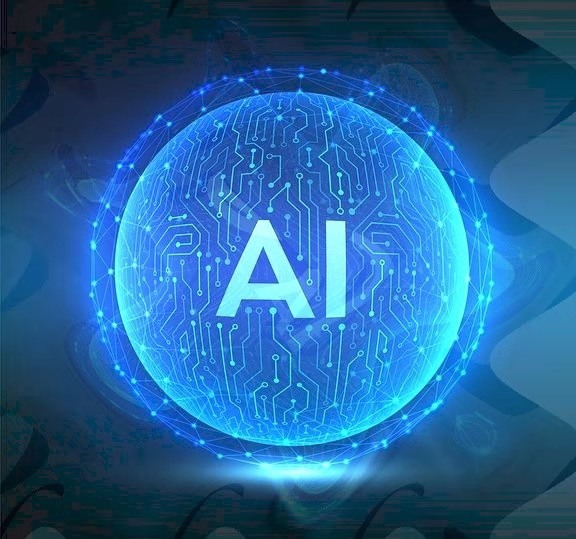}&
\hskip -0.2cm\includegraphics[clip, trim=0cm 1cm 0cm 0cm,width=0.33\columnwidth]{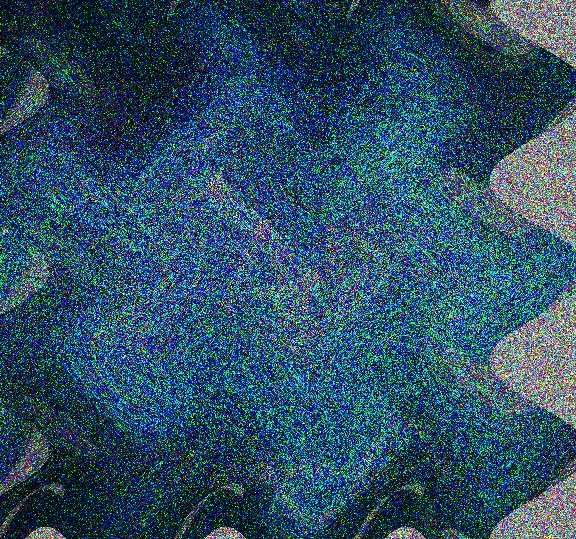}&
\hskip -0.2cm\includegraphics[clip, trim=0cm 1cm 0cm 0cm,width=0.33\columnwidth]{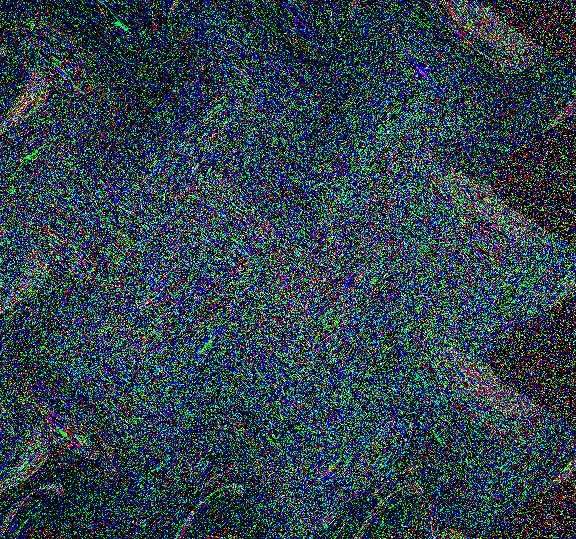}\\
\hskip -0.2cm{\small (a) $\vx(0)$ (Original)} &\hskip -0.2cm{\small (b) $\vx(1)$ (ODE)
}  & \hskip -0.2cm{\small (c) $\Tilde{\vx}(0)$ (ODE)
} & \hskip -0.2cm{\small (d) $\vx(1)$ (SDE)
}  & \hskip -0.2cm{\small (e) $\Tilde{\vx}(0)$ (SDE)
}\\
\end{tabular}
\vskip -0.2cm
\caption{\small Illustrations of the forward and backward propagation of the training data using 2D ODE in equation \eqref{Algorithm:Eq2} and SDE in equation \eqref{Algorithm:Eq4}.
(a) the original image; (b) and (d) the features of the original image generated by the forward propagation using ODE and SDE, respectively; (c) and (e) the recovered images by reverse-engineering the features shown in (b) and (d), respectively.  We see that it is easy to break the privacy of the ODE model but harder for SDE. 
}
\label{fig:ODE:vs:SDE-Privacy}\vspace{-0.4cm}
\end{figure*}

Given the training set ${S_N:=\{\vx_i, y_i\}_{i=1}^N}$, with $\{\vx_i, y_i\} \subset$ $\RR^d \times \RR$ being a data-label pair. For a given $\vx_i:=\vx^0$ the forward propagation of a ResNet with $M$ residual mappings can be written as
\begin{equation}\small
\label{Algorithm:Eq1}
\begin{aligned}
\vx^{l+1} &= \vx^l + \hat{F}(\vx^l, \mW^l), \ \ \mbox{for}\ l=0,\ldots, M-1,\\
\hat{y}_i &= f(\vx^{M}),
\end{aligned}
\end{equation}
where $\hat{F}(\cdot, \mW^l)$ is the nonlinear mapping of the $l$-th residual mapping parameterized by $\mW^l$; $f(\cdot)$ is the output activation, and $\hat{y}_i$ is the prediction. The heuristic continuous limit of \eqref{Algorithm:Eq1} is the following ordinary differential equation (ODE)
\begin{equation}\small
\label{Algorithm:Eq2}
d\vx(t) = F(\vx(t), \mW(t))dt,\ \ \vx(0) = \hat{\vx},
\end{equation}
where $t$ is the time variable. ODE \eqref{Algorithm:Eq2} can be revertible, and thus the ResNet counterpart might be exposed to data privacy leakage. For instance, we use an image downloaded from the Internet (Fig.~\ref{fig:ODE:vs:SDE-Privacy}(a)) as the initial data $\hat{\vx}$ in \eqref{Algorithm:Eq2}. Then we simulate the forward propagation of ResNet by solving \eqref{Algorithm:Eq2} from $t=0$ to $t=1$ using the forward Euler solver with a time step size $\Delta t= 0.01$ and a given velocity field $F(\vx(t), \mW(t))$ (see Sec.~\ref{sec:appendix:exp:details} for the details of $F(\vx(t), \mW(t))$), which maps the original image to its features (Fig.~\ref{fig:ODE:vs:SDE-Privacy}(b)). To recover the original image, we start from the feature and use the backward Euler iteration, i.e., 
$
\Tilde{\vx}(t) = \Tilde{\vx}(t+\Delta t) - \Delta tF(\tilde{\vx}(t+\Delta t), t+\Delta t),
$
to evolve $\tilde{\vx}(t)$ from $t=1$ to $t=0$ with $\Tilde{\vx}(1) = \vx(1)$ being the features obtained in the forward propagation. We plot the recovered image from features in Fig.~\ref{fig:ODE:vs:SDE-Privacy}(c), and the original image can be almost perfectly recovered.

\vspace{-0.4cm}
\subsection{Residual Perturbation and its SDE Analogue}\vspace{-0.1cm}
In this part, we propose two SDE models to reduce the reversibility of \eqref{Algorithm:Eq2}, and the corresponding residual perturbations analogue can protect the data privacy in DL. 

\paragraph{Strategy I} 
Consider the following SDE model:
\begin{gather}\label{Algorithm:Eq4}\small
d\vx(t) = F(\vx(t), \mW(t))dt + \gamma d\mB(t),\ \gamma >0,
\end{gather}
where $\mB(t)$ is the standard Brownian motion. {We simulate the forward propagation and reverse-engineering the input from the output} by solving the SDE model \eqref{Algorithm:Eq4} with $\gamma=1$ using the same $F(\vx(t), \mW(t))$ and initial data $\hat{\vx}$. We use the following forward \eqref{eq:forward:SDE1} and backward \eqref{eq:backward:SDE1} Euler-Maruyama discretizations  \cite{higham2001algorithmic} of \eqref{Algorithm:Eq4} for the forward and backward propagation, respectively.
\vspace{-0.1cm}
\begin{equation}\small
\label{eq:forward:SDE1}
\vx(t+\Delta t) = \vx(t) + \Delta t F(\vx(t), \mW(t)) + \gamma \mathcal{N}(\mathbf{0}, \sqrt{\Delta t}\,\mI),
\end{equation}
\begin{equation}\small
\label{eq:backward:SDE1}
\begin{aligned}
\tilde{\vx}(t) = \tilde{\vx}(t+\Delta t) &- \Delta tF(\tilde{\vx}(t+\Delta t), \mW(t+\Delta t))\\
&+ \gamma\mathcal{N}(\mathbf{0}, \sqrt{\Delta t}\,\mI),
\end{aligned}
\end{equation}
Fig.~\ref{fig:ODE:vs:SDE-Privacy}(d) and (e) show the results of the forward and backward propagation by SDE, respectively. These results show that it is much harder to reverse the features obtained from the SDE evolution. The SDE model informs us to inject Gaussian noise, in both training and test phases, to each residual mapping of ResNet to protect data privacy, which results in
\begin{equation}
\label{eq:ResidualPerturbation1}\small
\vx^{i+1} = \vx^i + \hat{F}(\vx^{i}, \mW^i) + \gamma \vn^{i}, \ \mbox{where}\ \vn^{i} \sim \mathcal{N}(\mathbf{0}, \mI) \footnote{
In \cite{liu2019neural,wang2019resnets}, the Gaussian noise injection has been used to improve the robustness of ResNets.}.
\end{equation}

\paragraph{Strategy II} For the second strategy, we consider using the multiplicative noise instead of the additive noise that used in \eqref{Algorithm:Eq4}  and \eqref{eq:ResidualPerturbation1}, and the corresponding SDE can be written as
\begin{gather}
\label{Algorithm:Eq3}\small
d\vx(t) = F(\vx(t), \mW(t))dt + \gamma \vx(t)\odot d\mB(t),\ \gamma >0,
\end{gather}
where $\odot$ denotes the Hadamard product. Similarly, we can use the forward and backward Euler-Maruyama discretizations of \eqref{Algorithm:Eq3} to propagate the image in Fig.~\ref{fig:ODE:vs:SDE-Privacy}(a), and we provide these results in 
Sec.~\ref{sec:appendix:multiplicative:noise}. The corresponding residual perturbation is 
\begin{equation}
\label{eq:ResidualPerturbation2}\small
\vx^{i+1} = \vx^i + \hat{F}(\vx^{i}, \mW^i) + \gamma \vx^i\odot \vn^{i}, \ \mbox{where}\ \vn^{i} \sim \mathcal{N}(\mathbf{0}, \mI).
\end{equation}
Again, the noise $\gamma \vx^i\odot \vn^{i}$ is injected to each residual mapping in both training and test stages.

We will provide theoretical guarantees for these two residual perturbation schemes, i.e., \eqref{eq:ResidualPerturbation1} and \eqref{eq:ResidualPerturbation2}, in Sec.~\ref{sec:Theory}, and numerically verify their efficacy in Sec.~\ref{sec:Experiments}.

\vspace{-0.4cm}
\subsection{Utility Enhancement via Model Ensemble}\vspace{-0.1cm}
In \cite{wang2019resnets}, the authors showed that an ensemble of noise-injected ResNets can improve models' utility. In this paper, we will also study the model ensemble for utility enhancement. We inherit notations from \cite{wang2019resnets}, e.g., we denote an ensemble of two noise-injected ResNet8 as En$_2$ResNet8.

\vspace{-0.3cm}
\section{Main Theory}\label{sec:Theory}

\vspace{-0.1cm}
\subsection{Differential Privacy Guarantee for Strategy I}
{We consider the following function class for ResNets with residual perturbation:
\begin{equation}\label{EnResNet1}\small
\mathcal F_1:=\left\{
f(\vx)=\vw^{\top}\vx^{M}|\vx^{i+1}=\vx^{i}+\phi\left(\mU^{i}\vx^{i}\right)+\gamma\vn^{i}\right\},
\end{equation}
where $\vx^0=\mbox{input data}+\pi\vn\in\RR^d$ is the noisy input with $\pi>0$ being a hyperparameter and $\vn\sim\mathcal N(\mathbf{0},\mI)$ being a normal random vector. $\mU^i\in\RR^{d\times d}$ is the weight matrix in the $i$-th residual mapping and $\vw\in\RR^d$ is the weights of the last layer. $\vn^i\sim\mathcal N(\mathbf{0},\mI)$ and $\gamma>0$ is also a hyperparameter. $\phi={\rm BN}(\psi)$ with ${\rm BN}$ being the batch normalization and $\psi$ being a $L$-Lipschitz and monotonically increasing activation function, e.g., ReLU.}

We first recap on the definition of differential privacy.
\begin{definition}[$(\epsilon,\delta)$-DP] \label{DP-Def-Dwork} \cite{dwork2006calibrating}
A randomized mechanism {\small $\cM:\cS^N\rightarrow\cR$} satisfies $(\epsilon,\delta)$-DP if for any two datasets {\small $S,S'\in \cS^N$} that differ by one element, and any output subset $O\subseteq \cR$, it holds that $\PP[\cM(S)\in O]\leq e^\epsilon\cdot \PP[\cM(S')\in O]+\delta, {\rm where}\,\,\,\delta\in(0,1)$ and $\epsilon>0.$
\end{definition}

Theorem~\ref{theorem1} below guarantees DP for {\bf Strategy I}, and we provide its proof in Sec.~\ref{sec:appendix:Thm1}.
\begin{theorem}\label{theorem1}
Assume the input to ResNet lies in $B(\mathbf{0}, R)$ and {the expectation} of the output of every residual mapping is normally distributed and bounded by {a constant} $G$, in $\ell_2$ norm.
Given the total number of iterations $T$ used for training ResNet. For any $\epsilon>0$ and $\delta,\lambda\in(0,1)$, the parameters $\mU^i$ and $\vw$ in the ResNet with residual perturbation satisfies {$\left((\lambda/(i+1)+(1-\lambda))\epsilon,\delta\right)$}-DP and  {$\left((\lambda/(M+1)+(1-\lambda))\epsilon,\delta\right)$}-DP, respectively, provided that $\pi>R\sqrt{(2Tb\alpha )/(N\lambda\epsilon)}$ and $\gamma>G\sqrt{(2Tb\alpha)/(N\lambda\epsilon)}$, where $\alpha=\log(1/\delta)/\left((1-\lambda)\epsilon\right)+1$, $M, N$ and $b$ are the number of residual mappings, training data, and batch size, respectively. {In particular, {in the above setting}
the whole model obtained by injecting noise according to strategy I satisfies $(\epsilon,\delta)$-DP}.
\end{theorem}

\vspace{-0.4cm}
\subsection{Theoretical Guarantees for Strategy II}

\paragraph{Privacy} 
To analyze the residual perturbation scheme in~\eqref{eq:ResidualPerturbation2}, we consider the following 
function class:
{\begin{equation}\label{EnResNet}\small
\hspace{-0.2cm}\mathcal{F}_2:=\left\{
f\left(\vx\right)=\vw^{\top}\vx^{M}+\pi \vx^{M}\vn|\Bigg|
\begin{aligned}
    &\vx^{i+1}=\vx^{i}+\phi\left(\mU^{i}\vx^{i}\right)+\\ \nonumber
    &\gamma \tilde{\vx}^{i}\odot\vn^i,i=0,\ldots,M-1
\end{aligned}    
\right\}
\end{equation}
where $\vn^i\sim\mathcal{N}(\mathbf{0},\mI)$ is a normal random vector for $i=0,\ldots,M-1$, $\|\vw\|\leq a$ with $a>0$ being a constant.} We denote the entry of $\vx^i$ that has the largest absolute value as $x^i_{max}$, and $\tilde{\vx}^i := (sgn(x^i_{j})\max(|x^i_{j}|,\eta))_{j=1}^d$. Due to batch normalization, we assume {the $l^2$ norm of} $\phi$ can be bounded by a positive constant $B$. The other notations are defined similar to that in \eqref{EnResNet1}. Consider training $\mathcal{F}_2$ by using two datasets $S$ and $S'$, and we denote the resulting models as:
\begin{equation}\label{eq:models:two-datasets1}\small
\begin{aligned}
 \hspace{-0.5cm}    f\left(\vx|S\right)&:=\vw_{1}^{\top}\vx^{M}+\pi \vx^{M}\odot\vn^M;\,\,\ \mbox{where}\\
    \vx^{i+1}&=\vx^{i}+\phi\left(\mU_{1}^{i}\vx^{i}\right)+\gamma \tilde{\vx}^{i}\odot\vn^i,i=0,\ldots,M-1,
\end{aligned}
\end{equation}
and
\begin{equation}\label{eq:models:two-datasets2}\small
\begin{aligned}
\hspace{-0.4cm}f\left(\vx|S^{'}\right)&:=\vw_{2}^{\top}\vx^{M}+\pi \vx^{M}\odot\vn^M;\,\, \ \mbox{where}\\
    \vx^{i+1}&=\vx^{i}+\phi\left(\mU_{2}^{i}\vx^{i}\right)+\gamma \tilde{\vx}^{i}\odot\vn^{i},i=0,\ldots,M-1.
\end{aligned}
\end{equation}
Then we have the following privacy guarantee.
\begin{theorem}\label{thm:dp}
For $f(\vx|S)$ and $f(\vx|S')$ that are defined in \eqref{eq:models:two-datasets1} and \eqref{eq:models:two-datasets2}, respectively. Let $\lambda\in(0,1),\delta\in(0,1)$, and $\epsilon>0$, if
${\small \\ \gamma>(B/\eta)\sqrt{(2\alpha M)/(\lambda\epsilon)}}\ \ \mbox{and}\ \ {\small \pi>a\sqrt{\left(2\alpha M\right)/\lambda\epsilon},}$ where $\alpha=\log(1/\delta)/\left((1-\lambda)\epsilon\right)+1$, then $\PP[f(\vx|S)\in O]\leq e^\epsilon\cdot \PP[f(\vx|S^{'}]\in O]+\delta$ for any input $\vx$ and any subset $O$ in the output space. 
\end{theorem}
We prove Theorem~\ref{thm:dp} in Sec.~\ref{sec:appendix:Thm2}. Theorem~\ref{thm:dp} guarantees the privacy of the training data given only black-box access to the model, i.e., the model will output the prediction for any input without granting adversaries access to the model itself. In particular, we cannot infer whether the model is trained on $S$ or $S'$ no matter how we  query the model in a black-box fashion. This kind of privacy-preserving prediction has also been studied in \cite{Papernot:2017,Papernot:2018,dwork2018privacy,bassily2018model}. {We leave theoretical DP-guarantee for for Strategy II for future work.}

\paragraph{Generalization gap} Many works have shown that overfitting in training ML models leads to privacy leakage \cite{salem2018ml}, and reducing overfitting can mitigate data privacy leakage~\cite{shokri2017membership,yeom2018privacy,sablayrolles2018d,salem2018ml,wu2019p3sgd}. In this part, we show that the residual perturbation \eqref{eq:ResidualPerturbation2} can reduce overfitting via computing the Rademacher complexity. For simplicity, we consider binary classification problems. Suppose {\small$S_N=$}{\small $\{\vx_i, y_i\}_{i=1}^N$} is drawn from {\small $X \times Y \subset \RR^d \times \{-1,+1\}$} with $X$ and $Y$ being the input data and label spaces, respectively. Assume $\mathcal{D}$ is the underlying distribution of {\small $X \times Y$}, which is unknown.  Let {\small $\mathcal{H}\subset V$} be the hypothesis class of the ML model. We first recap on the definition of Rademacher complexity.

\begin{definition}\label{def:Rademacher:Complexity}
\cite{peter2002rademacher} Let $\mathcal H:X\rightarrow\RR$ be the space of real-valued functions on the space $X$. For a given sample $S=\{\vx_1,\vx_2,\ldots,\vx_N\}$ of size $N$, the empirical Rademacher complexity of $\mathcal H$ is defined as
\vspace{-0.15cm}
\begin{equation*}\small
		R_S(\mathcal H) := \frac{1}{N}E_{\sigma}\Big[\sup_{h\in \mathcal H}\sum_{i=1}^N\sigma_ih(\vx_i)\Big],
\end{equation*}
where $\sigma_1,\ldots,\sigma_N$ are i.i.d. Rademacher random variables with $\mathbb{P}(\sigma_i=1)=\mathbb{P}(\sigma_i=-1)=1/2$.
\end{definition}
Rademacher complexity is a tool to bound the generalization gap \cite{peter2002rademacher}.
The smaller the generalization gap is, the less overfitting the model is. For $\forall \vx_i\in \RR^d$ and constant $c\geq 0$, we consider the following two function classes ($\mathcal{F}$ and $\mathcal{G}$):
{\begin{equation*}\small
\begin{aligned}
\mathcal{F} := 
\left\{f(\vx, \vw)=\vw\vx^p(T)\big| d\vx(t)=\mU\vx(t)dt\ \text{with}\ \vx(0) = \vx_i\right\}  
\end{aligned},
\end{equation*}
where $\vw\in \RR^{1\times d},\mU\in \RR^{d\times d}\ \mbox{with}\ \|\vw\|_2\ \text{and}\ \|\mU\|_2\leq c$.} And 
{\begin{equation*}\small
\begin{aligned}
\mathcal{G}:=\big\{
f(\vx,\vw)=\mathbb E\left[\vw\vx^p(T)\right]\big|
d\vx(t)&=\mU\vx(t)dt\\
&+\gamma\vx(t)\odot dB(t)
\big\},
\end{aligned}
\end{equation*}
where $\vx(0)=\vx_i$, $\vw\in\RR^{1\times d}$ and $\mU\in\RR^{d\times d}$ with $\|\vw\|_2,\|\mU\|_2\leq c$,} $0<p<1$ takes the value such that $\vx^p$ is well defined on 
$\RR^d$, $\gamma>0$ is a hyperparameter and $\mU$ is a circulant matrix that corresponding to the convolution layer in DNs, and $B(t)$ is the standard 1D Brownian motion. The function class $\mathcal{F}$ represents the continuous analogue of ResNet without inner nonlinear activation functions, and $\mathcal{G}$ denotes $\mathcal{F}$ with the residual perturbation~\eqref{eq:ResidualPerturbation2}.

\begin{theorem}
\label{Theory:Thm1}
Given the training set $S_N=\{\vx_i,y_i\}_{i=1}^N$. We have $R_{S_N}(\mathcal{G}) < R_{S_N}(\mathcal{F})$.
\end{theorem}

We provide the proof of Theorem~\ref{Theory:Thm1} in Sec.~\ref{sec:appendix:Thm3}, where we will also provide quantitative lower and upper bounds of the above Rademacher complexities.  Theorem~\ref{Theory:Thm1} shows that residual perturbation \eqref{eq:ResidualPerturbation2} can reduce the generalization error. We will numerically verify this generalization error reduction for ResNet with residual perturbation in Sec.~\ref{sec:Experiments}.

\vspace{-0.2cm}
\section{Experiments}\label{sec:Experiments}
{In this section, we will verify that 1) Residual perturbation can protect data privacy, particularly membership privacy. 2) The ensemble of ResNets with residual perturbation improves the classification accuracy. 3) The skip connection is crucial in residual perturbation for DL with data privacy. 4) Residual perturbation can improve the utility of the resulting ML models over DPSGD with a compromise in privacy protection.}
We focus on {\small \bf Strategy I} in this section, and we provide the results of {\small \bf Strategy II} in Sec.~\ref{sec:appendix:strategyII}.

\vspace{-0.4cm}
\subsection{Preliminaries}
\begin{figure}[t!]
\centering
\begin{tabular}{cccc}
\hskip -0.6cm\includegraphics[clip, trim=0cm 0cm 0cm 2cm,width=0.31\columnwidth]{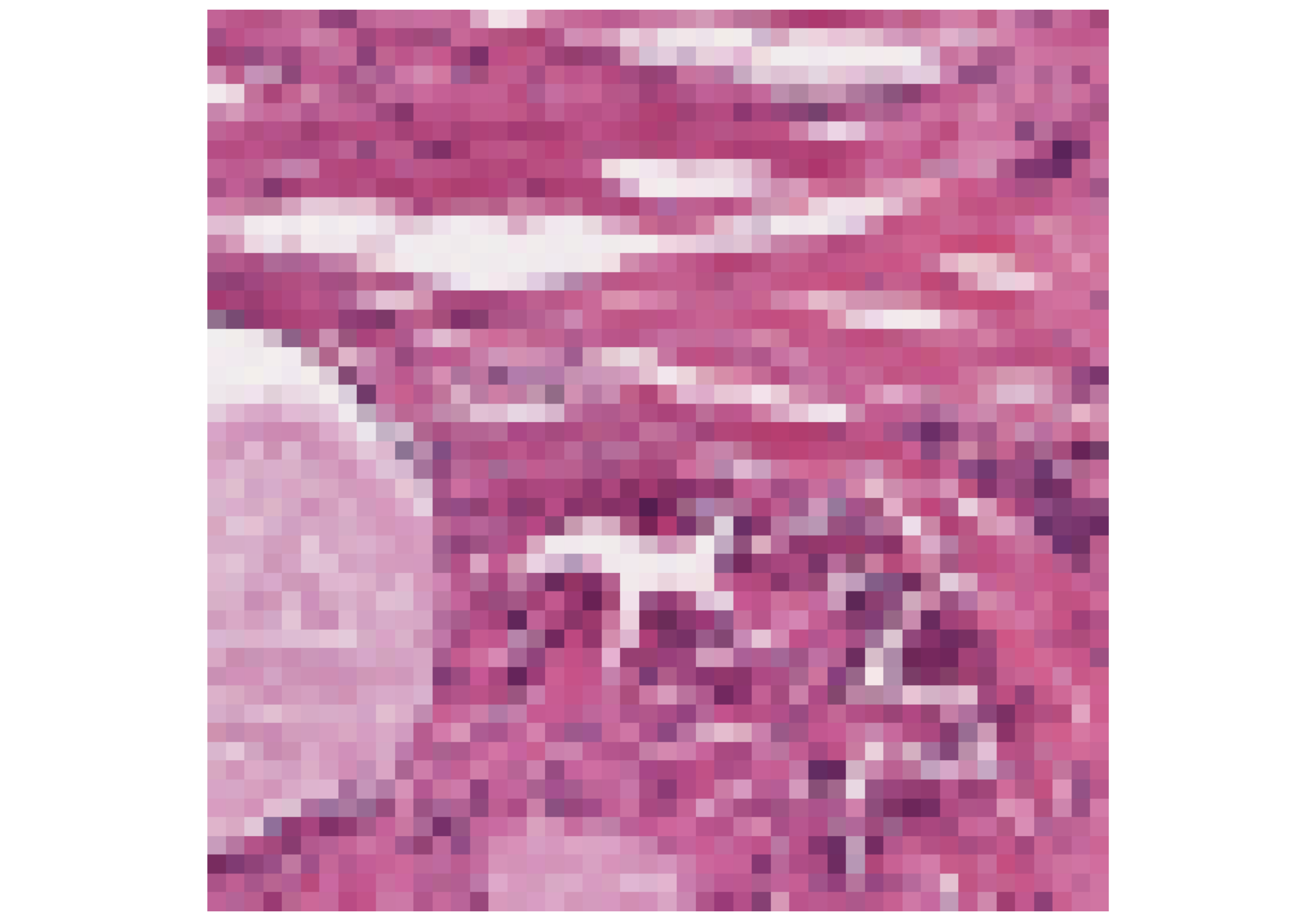}&
\hskip -0.9cm\includegraphics[clip, trim=0cm 0cm 0cm 2cm,width=0.31\columnwidth]{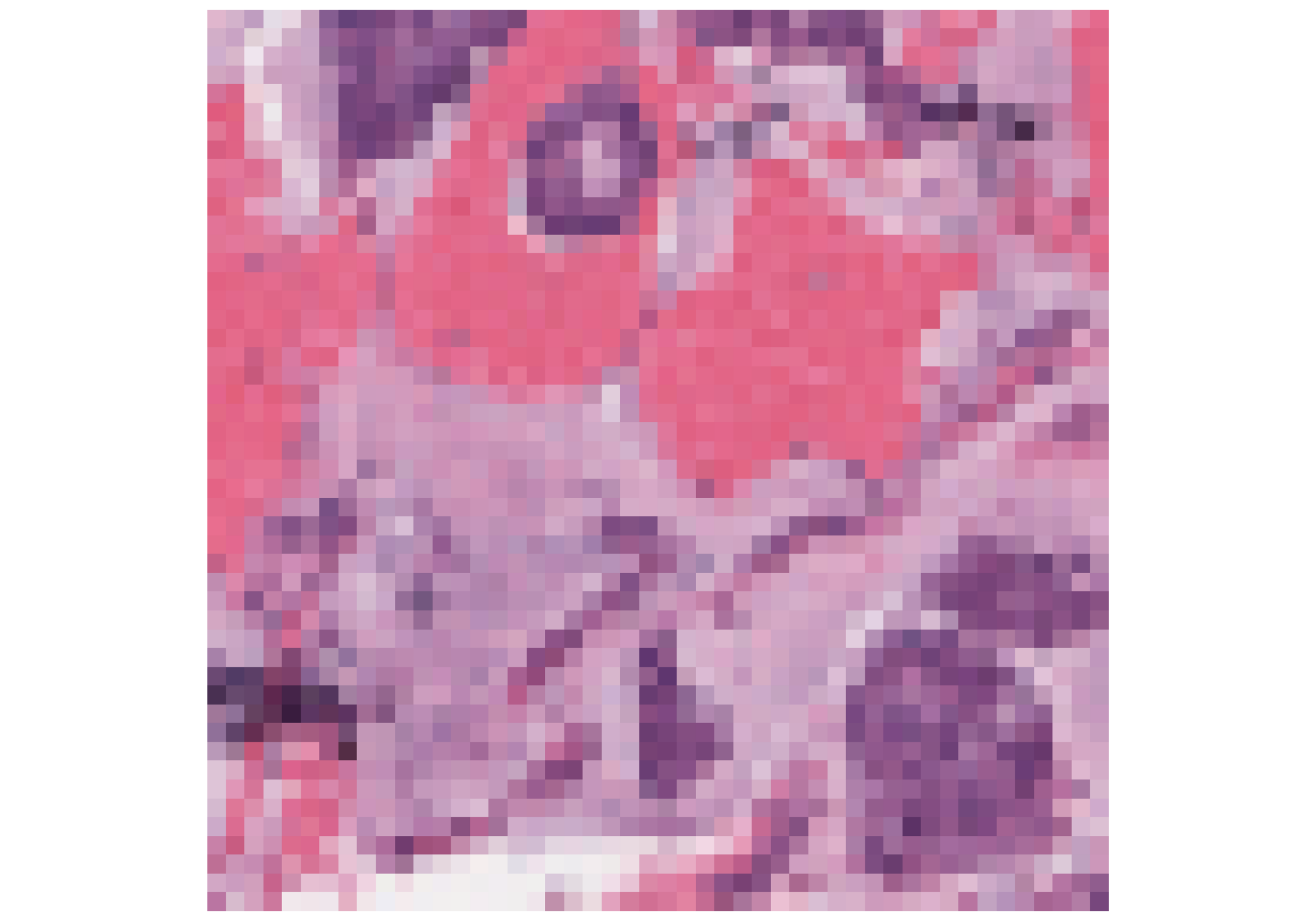}& 
\hskip -0.9cm\includegraphics[clip, trim=0cm 0cm 0cm 2cm,width=0.31\columnwidth]{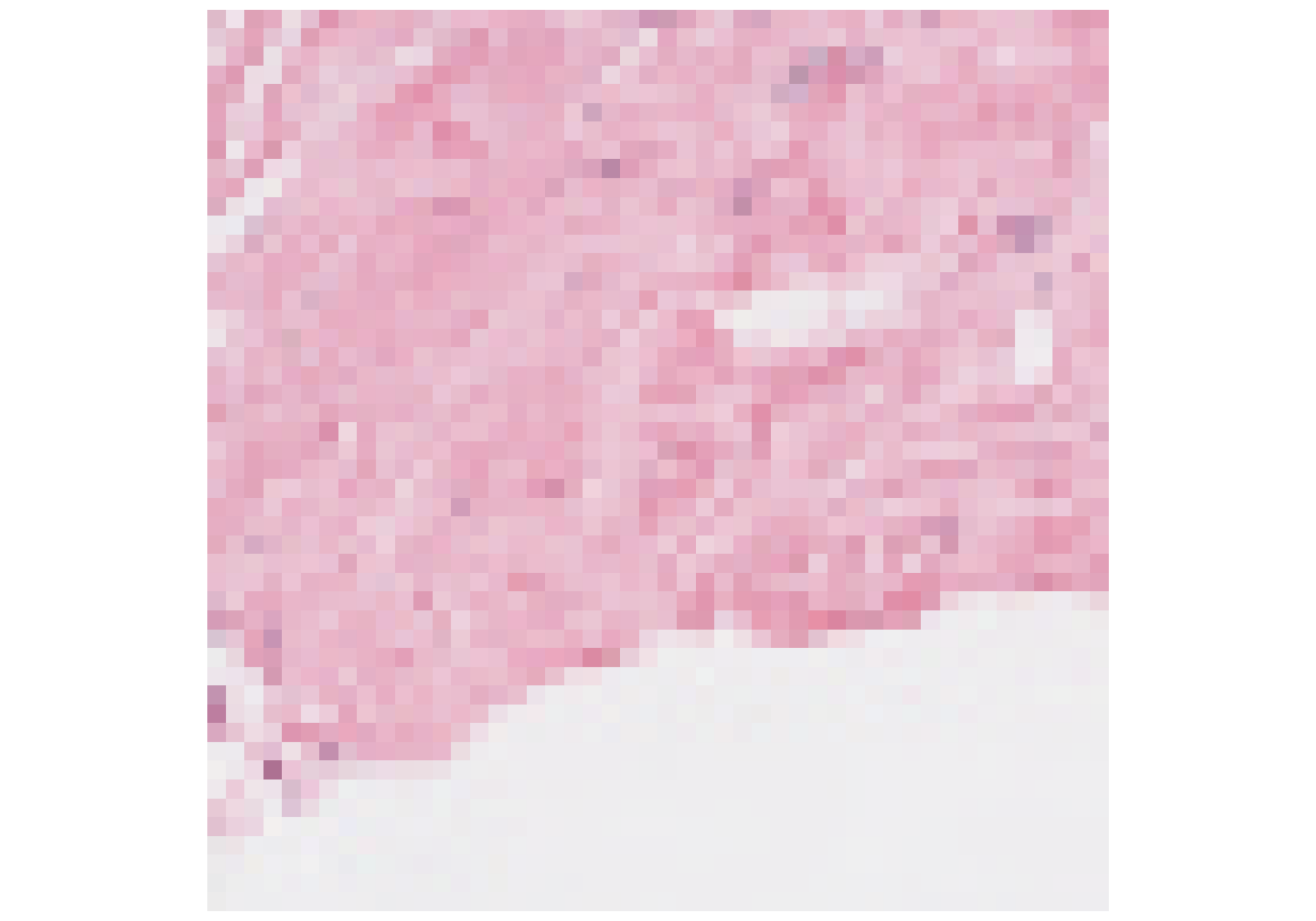}&
\hskip -0.9cm\includegraphics[clip, trim=0cm 0cm 0cm 2cm,width=0.31\columnwidth]{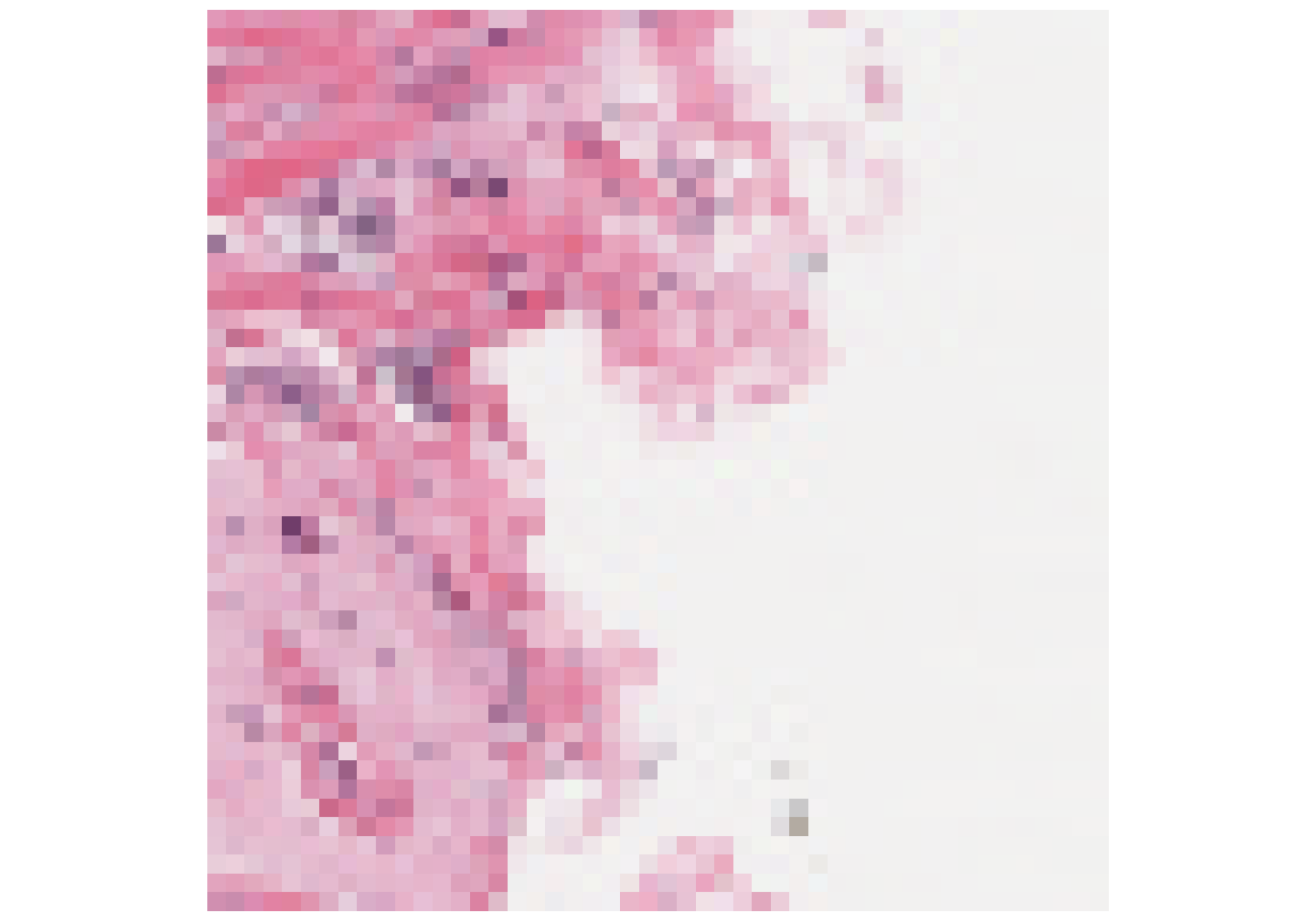}\\
\hskip -0.6cm {\small Negative} & \hskip -0.9cm {\small Positive} & \hskip -0.9cm {\small Negative} & \hskip -0.9cm {\small Positive} \\
\end{tabular}
\vskip -0.2cm
\caption{\small Visualization of a few selected images from the IDC dataset.}
\label{fig:IDC_visual2}\vspace{-0.1cm}
\end{figure}

\paragraph{Datasets} We consider {MNIST}, CIFAR10/CIFAR100 \cite{krizhevsky2009learning}, and the Invasive Ductal Carcinoma \cite{IDC:dataset} datasets. {MNIST consists of 70K $28\times 28$ gray-scale images with 60K and 10K of them used for training and test, respectively.} Both CIFAR10 and CIFAR100 contain 60K $32 \times 32$ color images with 50K and 10K of them used for training and test, respectively. 
The IDC dataset is a breast cancer-related 
benchmark dataset, which contains 277,524 patches of the size $50 \times 50$ with 198,738 labeled negative (0) and 78,786 labeled positive (1). Fig.~\ref{fig:IDC_visual2} depicts a few patches from the IDC dataset. For the IDC dataset, we follow~\cite{wu2019characterizing} and split the whole dataset into training, validation, and test set. The training set consists of 10,788 positive patches and 29,164 negative patches, and the test set contains 11,595 positive patches and 31,825 negative patches. The remaining patches are used as the validation set. For each dataset, we split its training set into $D_{\rm shadow}$ and $D_{\rm target}$ with the same size. Furthermore, we split $D_{\rm shadow}$ into two halves with the same size and denote them as $D_{\rm shadow}^{\rm train}$ and $D_{\rm shadow}^{\rm out}$, and split $D_{\rm target}$ by half into $D_{\rm target}^{\rm train}$ and $D_{\rm target}^{\rm out}$. The purpose of this splitting of the training set is for the membership inference attack, which will be discussed below.

\paragraph{Membership inference attack} To verify the efficiency of residual perturbation for privacy protection, we consider the  membership inference attack \cite{salem2018ml} in all the experiments below. The membership attack proceeds as follows: 1) train the shadow model by using $D_{\rm shadow}^{\rm train}$; 2) apply the trained shadow model to predict all data points in $D_{\rm shadow}$ and obtain the corresponding classification probabilities of belonging to each class. Then we take the top three classification probabilities (or two for 
binary classification) to form the feature vector for each data point. A feature vector is tagged as $1$ if the corresponding data point is in $D_{\rm shadow}^{\rm train}$, and $0$ otherwise. Then we train the attack model by leveraging all the labeled feature vectors; 3) train the target model by using $D_{\rm target}^{\rm train}$ and obtain feature vector for each point in $D_{\rm target}$. Finally, we leverage the attack model to decide whether a data point is in $D_{\rm target}^{\rm train}$.

\paragraph{Experimental settings} 
We consider En$_5$ResNet8 (ensemble of 5 ResNet8 with residual perturbation) and the standard ResNet8 as the target and shadow models, respectively. We use a multilayer perceptron with a hidden layer of $64$ nodes, followed by a softmax function as the attack model, which is adapted from \cite{salem2018ml}. We apply the same settings in \cite{he2016deep} to train the target and shadow models on MNIST, CIFAR10, and CIFAR100. For training models on the IDC dataset, we run 100 epochs of SGD with the same setting as before except that we decay the learning rate by 4 at the 20th, 40th, 60th, and 80th epoch, respectively. Moreover, we run $50$ epochs of Adam \cite{kingma2014adam} with a learning rate of 0.1 to train the attack model. For both IDC, MNIST, and CIFAR datasets, we set $\pi$ as half of $\gamma$ in {\bf Strategy I}, which simplifies hyperparameters calibration, and based on our experiment it gives a good tradeoff between privacy and accuracy. 

\paragraph{Performance evaluations} 
We consider both classification accuracy and capability for protecting membership privacy. The attack model is a binary classifier, which is used to decide if a data point is in the training set of the target model. For any $\vx \in D_{\rm target}$, we apply the attack model to predict its probability ($p$) of belonging to the training set of the target model. Given any fixed threshold $t$ if $p\geq t$, we classify $\vx$ as a member of the training set (positive sample), and if $p < t$, we conclude that $\vx$ is not in the training set (negative sample); so we can obtain different attack results with different thresholds. Furthermore, we can plot the ROC curve(see details in Sec.~\ref{subsection:ROC}) of the attack model and use the area under the ROC curve (AUC) as an evaluation of the membership inference attack. {The target model protects perfect membership privacy if the AUC is 0.5 (the attack model performs a random guess), and the more AUC deviates from 0.5, the less private the target model is.} Moreover, we use the precision (the fraction of records inferred as members are indeed members of the training set) and recall (the fraction of training set that is correctly inferred as members of the training set by the attack model) to measure ResNets' capability for protecting membership privacy.

\begin{figure}[!ht]
\centering
\begin{tabular}{cc}
\hskip -0.45cm\includegraphics[clip, trim=0cm 0cm 0cm 0.8cm,width=0.48\columnwidth]{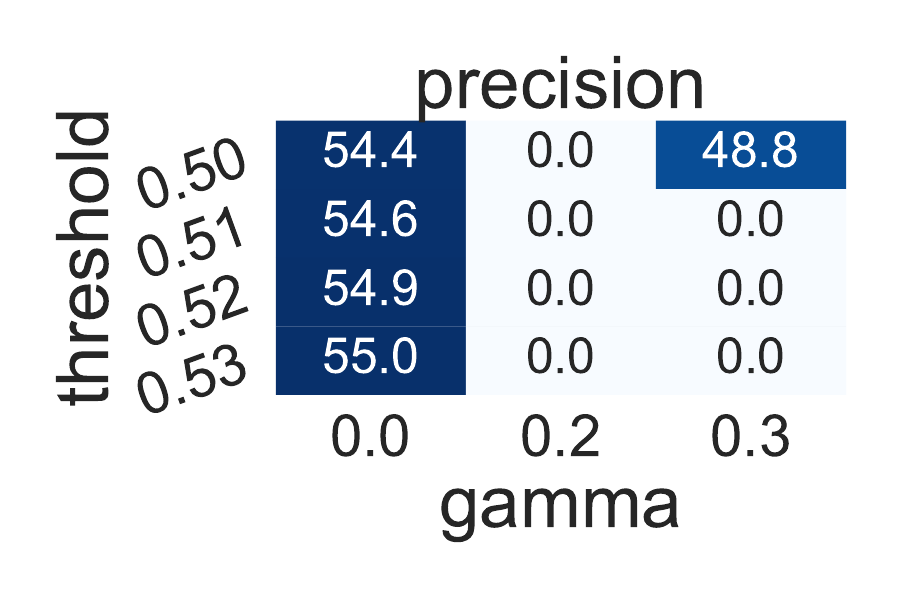}&
\hskip -0.45cm\includegraphics[clip, trim=0cm 0cm 0cm 0.8cm,width=0.48\columnwidth]{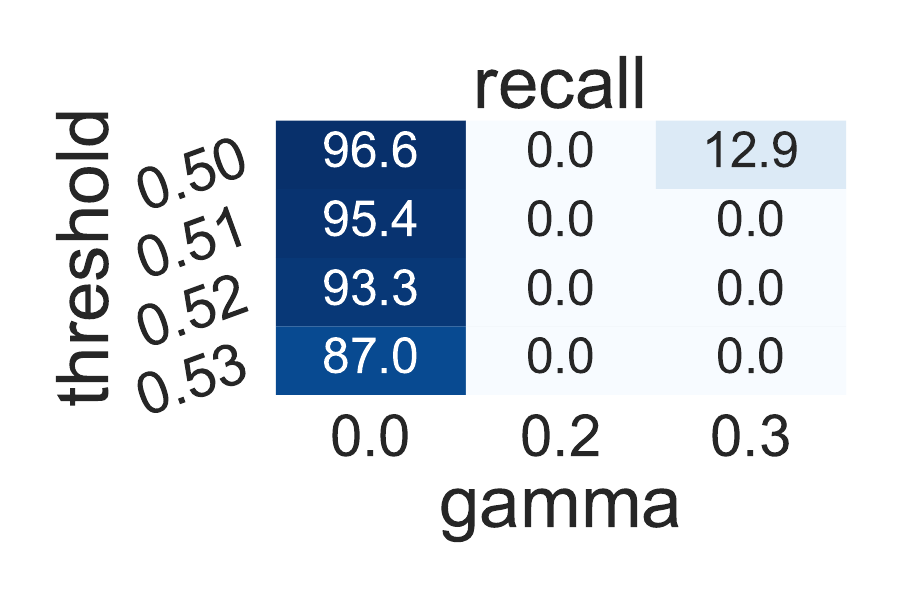}\\
\hskip -0.45cm\includegraphics[clip, trim=0cm 0cm 0cm 0.8cm,width=0.48\columnwidth]{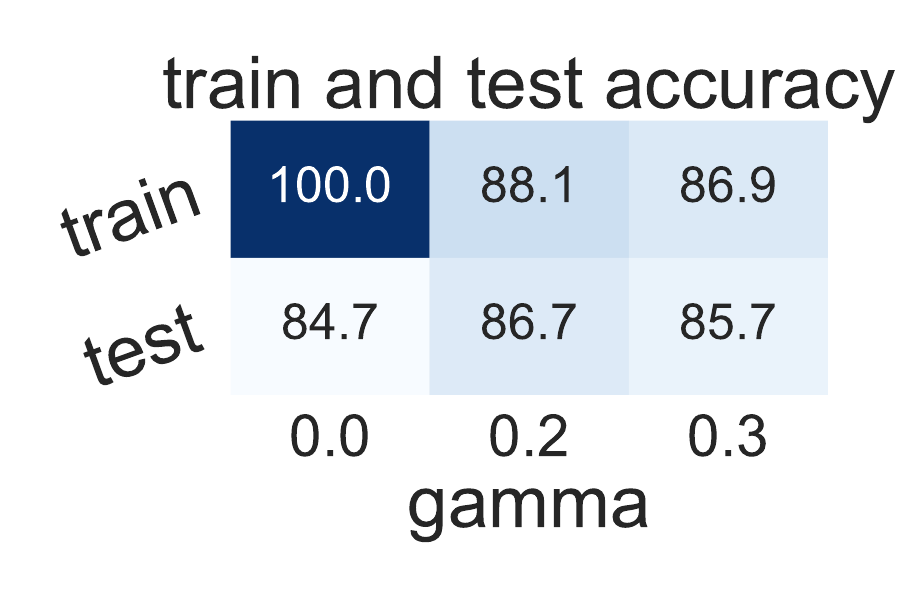}&
\hskip -0.45cm\includegraphics[clip, trim=0cm 0cm 0cm 0.8cm,width=0.48\columnwidth]{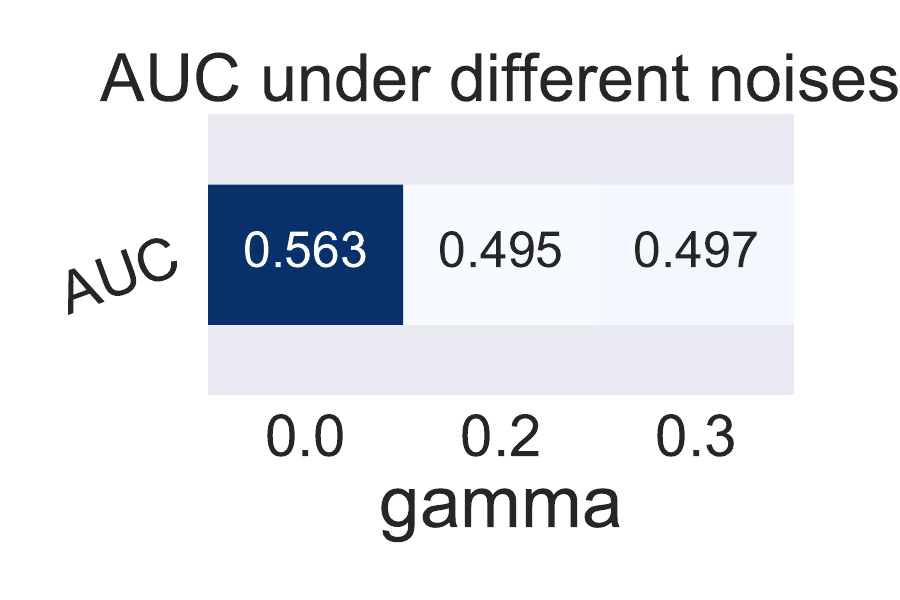}\\
\end{tabular}
\vskip -0.4cm
\caption{\small Performance of residual perturbation {(\textbf{Strategy I})} for En$_5$ResNet8 with different noise coefficients ($\gamma$) and membership inference attack thresholds on the IDC dataset. Residual perturbation 
significantly 
improves membership privacy and reduces the generalization gap. $\gamma=0$ corresponding to the baseline ResNet8. (Unit: \%)}
\label{fig:IDC_cn}\vspace{-0.4cm}
\end{figure}

\vspace{-0.4cm}
\subsection{Experiments on the IDC Dataset}
{In this subsection, we numerically verify that residual perturbation protects data privacy while retaining the classification accuracy of the IDC dataset.} We select the En$_5$ResNet8 as a benchmark architecture, which has ResNet8 as its baseline architecture. As shown in Fig.~\ref{fig:IDC_cn}, we set four different thresholds to obtain different attack results with three different noise coefficients ($\gamma$), and $\gamma=0$ means the standard ResNet8 without residual perturbation. We also depict the ROC curve for this experiment in Fig.~\ref{fig:ROC_cn}(c). En$_5$ResNet8 is remarkably better in protecting the membership privacy, and as $\gamma$ increases the model becomes more resistant to the membership attack. Also, as the noise coefficient increases, the gap between training and test accuracies becomes smaller, which resonates with Theorem~\ref{Theory:Thm1}. For instance, when $\gamma=0.2$ the AUC for the attack model is $0.495$ and $0.563$, respectively, for En$_5$ResNet8 and ResNet8; the classification accuracy of En$_5$ResNet8 and ResNet8 are 0.867 and 0.847, respectively.

\paragraph{Residual perturbation vs. DPSGD}
In this part, we compare the residual perturbation with the benchmark Tensorflow DPSGD module \cite{mcmahan2018general}, and we calibrate the hyperparameters, including the initial learning rate (0.1) which decays by a factor of 4 after every 20 epochs, noise multiplier (1.1), clipping threshold (1.0), micro-batches (128), and epochs (100) \footnote{\url{https://github.com/tensorflow/privacy/tree/master/tutorials}} such that the resulting model gives the optimal tradeoff between membership privacy and classification accuracy. {DPSGD is significantly more expensive due to the requirement of computing and clipping the per-sample gradient.} We compare the standard ResNet8 trained by DPSGD with the baseline non-private ResNet8, and  En$_5$ResNet8 with residual perturbation ($\gamma=0.5$). Table~\ref{tab:RP:vs:DPSGD_cn} lists the AUC of the attack model and training and test accuracies of the target model; {we see that DPSGD and residual perturbation can protect membership privacy, and residual perturbation can improve accuracy without sacrificing membership privacy protection.}

\begin{table}[!ht]
\caption{\small Residual perturbation {(\textbf{Strategy I})} vs. DPSGD in training ResNet8 and EnResNet8 for IDC classification. 
Residual perturbation has 
better accuracy and protects better membership privacy than the baseline ResNet8 (AUC closer to 0.5).}\label{tab:RP:vs:DPSGD_cn}
\centering
\fontsize{9.0pt}{0.9em}\selectfont
\begin{tabular}{c|c|c|c}
\hline
Model  & AUC & Training Acc & Test Acc\\
\hline
{ResNet8(non-private)}   & {0.563} & {1.000} & {0.847}\\
ResNet8 (DPSGD) & \bf 0.503 & 0.831 & 0.828\\
{En$_1$ResNet8($\gamma=0.5$)}   & \bf {0.499} & \bf {0.882} & \bf {0.865}\\
\hline
\end{tabular}
\end{table}

\vspace{-0.4cm}
\subsection{Experiments on the MNIST/CIFAR10/CIFAR100 Datasets}
\begin{figure}[!ht]
\centering
\begin{tabular}{cc}
\hskip -0.45cm\includegraphics[clip, trim=0cm 0cm 0cm 0.8cm,width=0.48\columnwidth]{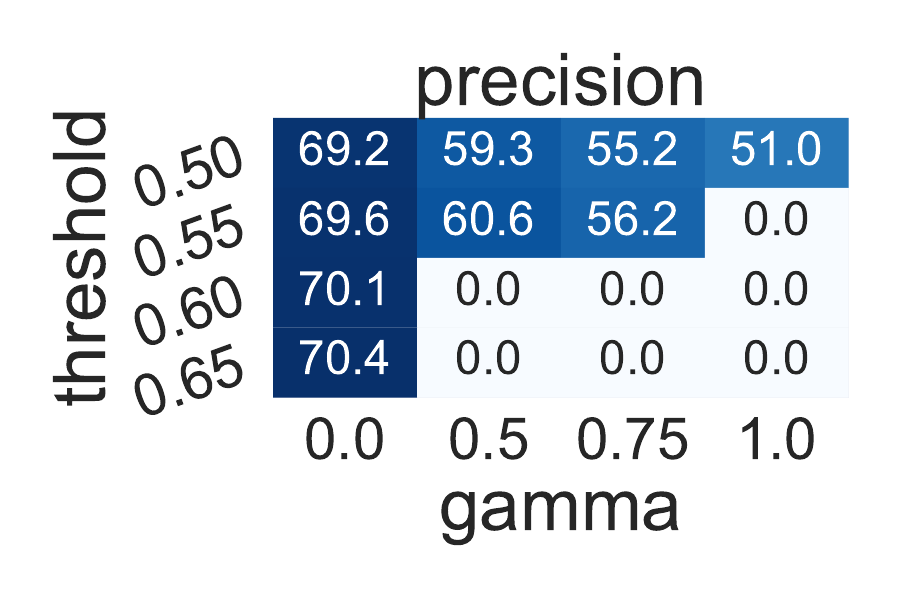}&
\hskip -0.45cm\includegraphics[clip, trim=0cm 0cm 0cm 0.8cm,width=0.48\columnwidth]{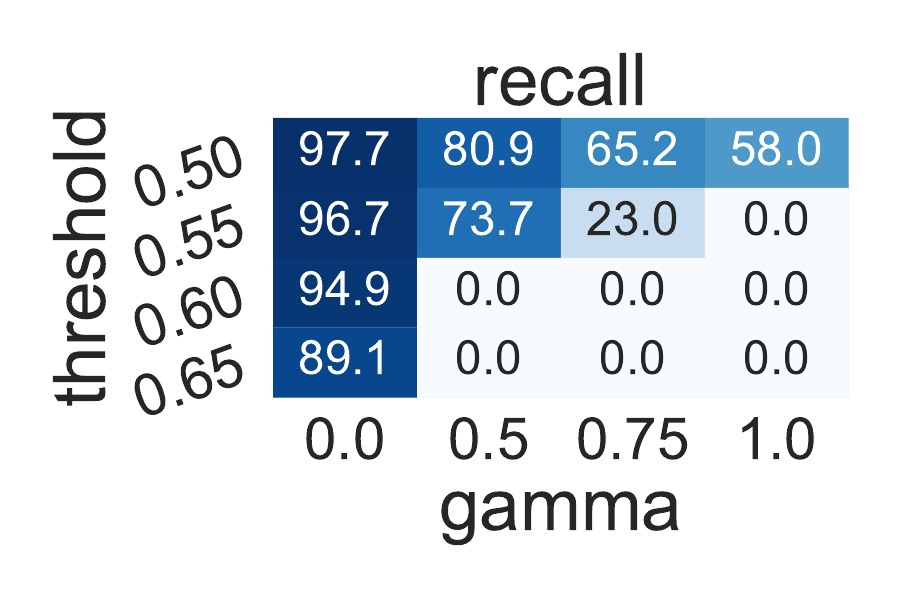}\\
\hskip -0.45cm\includegraphics[clip, trim=0cm 0cm 0cm 0.8cm,width=0.48\columnwidth]{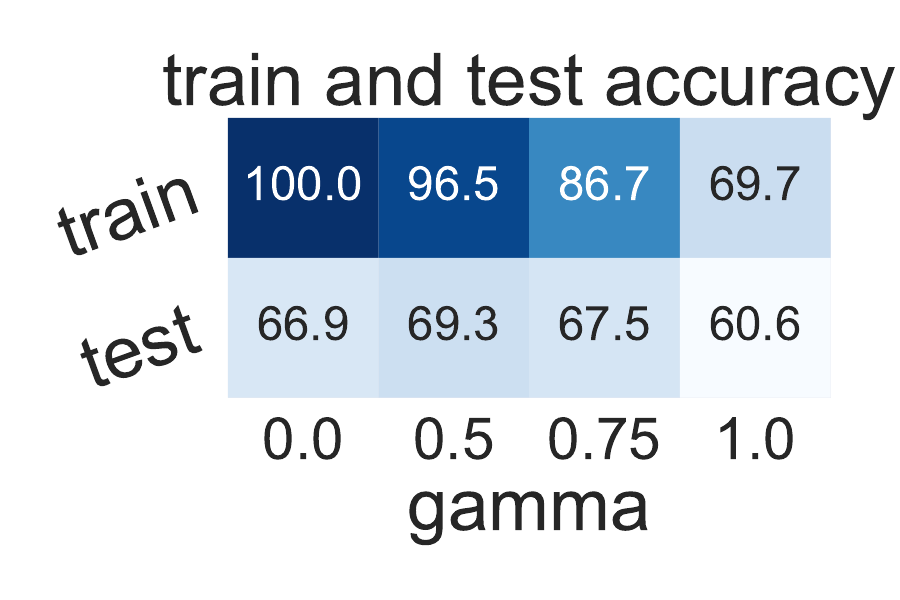}&
\hskip -0.45cm\includegraphics[clip, trim=0cm 0cm 0cm 0.8cm,width=0.48\columnwidth]{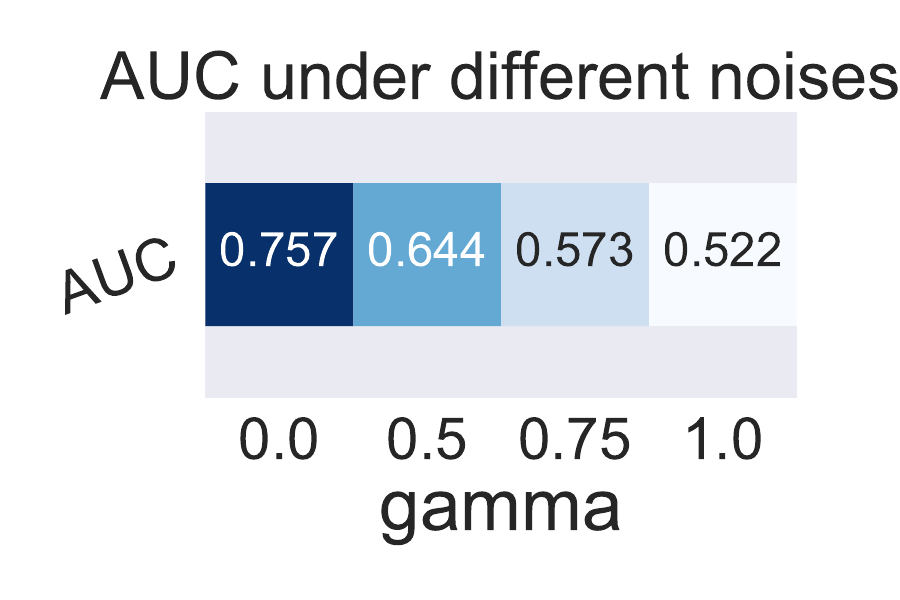}\\ 
\end{tabular}
\vskip -0.4cm
\caption{\small 
Performance of En$_5$ResNet8 with residual perturbation {(\textbf{Strategy I})} using different noise coefficients ($\gamma$) and membership inference attack threshold on CIFAR10. 
Residual perturbation can not only enhance the membership privacy, but also improve the classification accuracy. 
$\gamma=0$ corresponding to the baseline ResNet8 without residual perturbation or model ensemble. (Unit: \%)}
\label{fig:Cifar10-WithSkip_cn}\vspace{-0.4cm}
\end{figure}

\begin{figure}[!ht]
\centering
\begin{tabular}{cc}
\hskip -0.45cm\includegraphics[clip, trim=0cm 0cm 0cm 0cm,width=0.48\columnwidth]{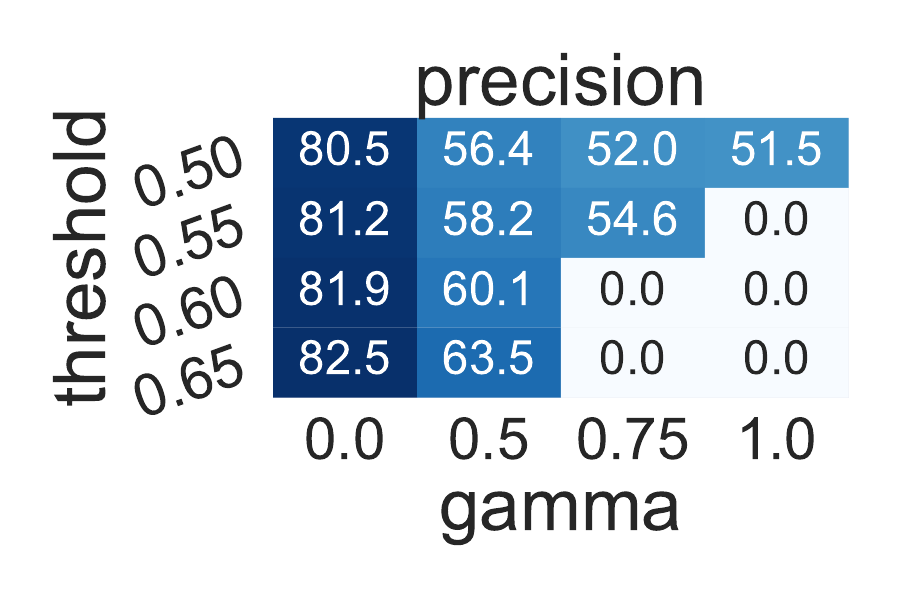}&
\hskip -0.45cm\includegraphics[clip, trim=0cm 0cm 0cm 0cm,width=0.48\columnwidth]{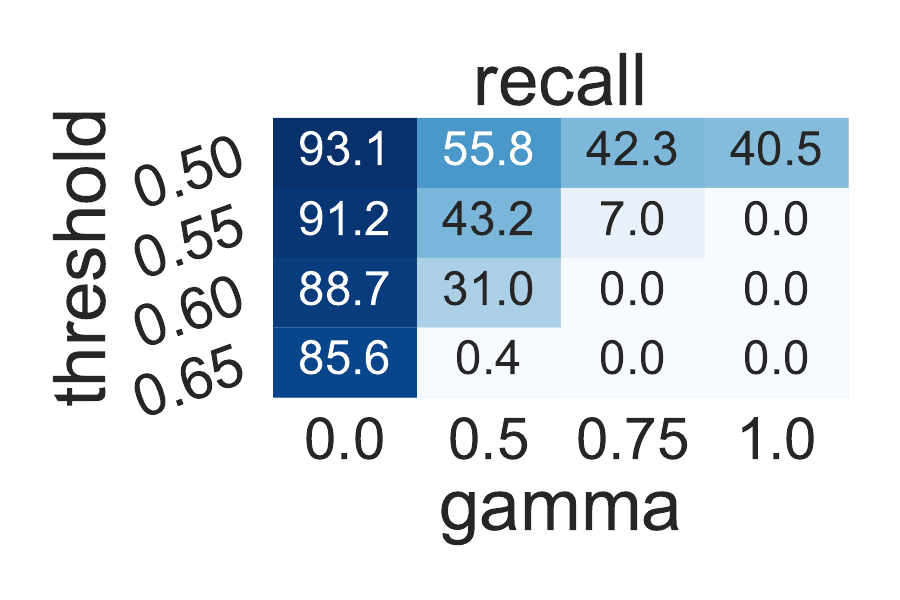}\\
\hskip -0.45cm\includegraphics[clip, trim=0cm 0cm 0cm 0cm,width=0.48\columnwidth]{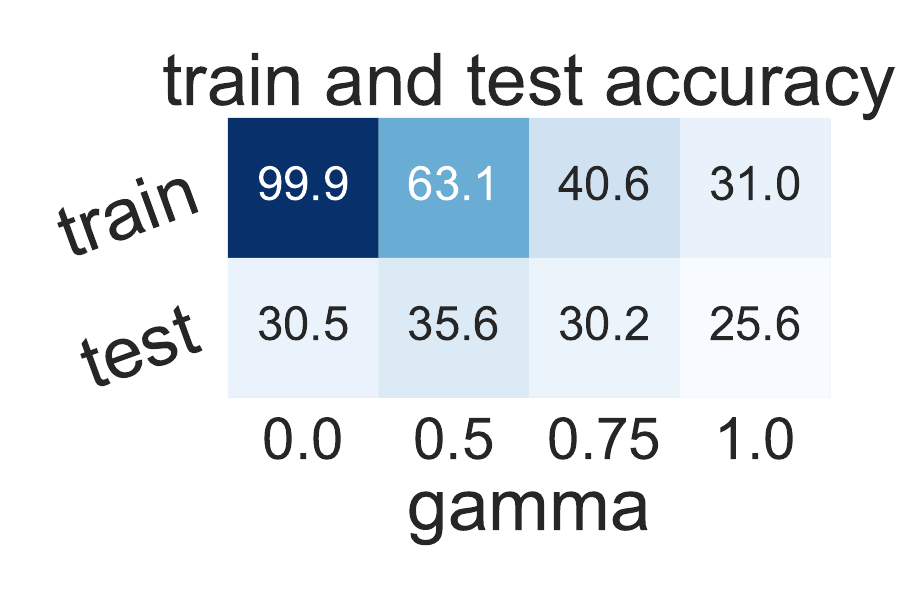}&
\hskip -0.45cm\includegraphics[clip, trim=0cm 0cm 0cm 0cm,width=0.48\columnwidth]{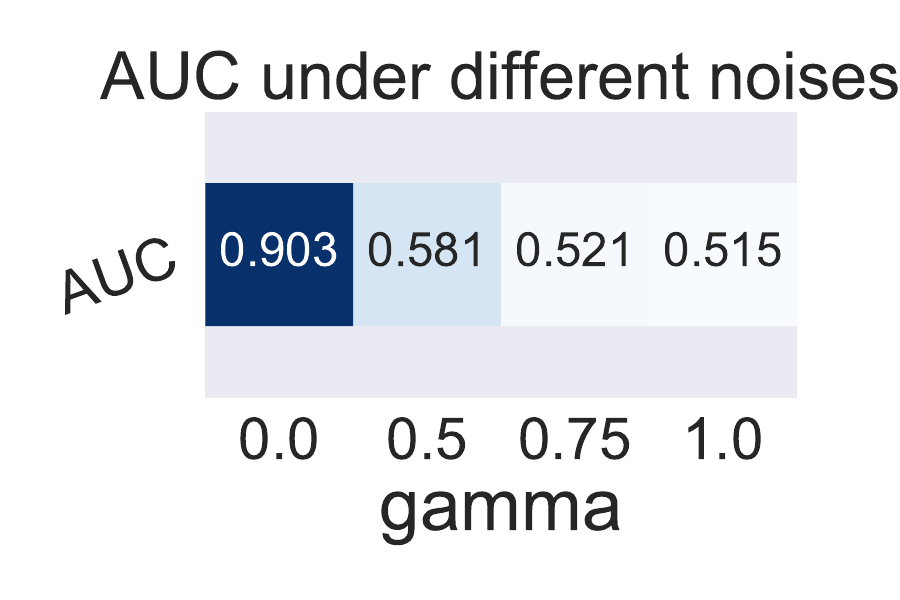}\\
\end{tabular}
\vskip -0.2cm
\caption{ \small 
Performance of En$_5$ResNet8 with residual perturbation {(\textbf{Strategy I})} using different noise coefficients ($\gamma$) and membership inference attack threshold on CIFAR100. Again, residual perturbation can not only enhance the membership privacy, but also improve the classification accuracy. 
$\gamma=0$ corresponding to the baseline ResNet8 without residual perturbation or model ensemble.
(Unit: \%)
}
\label{fig:Cifar100_cn}\vspace{-0.2cm}
\end{figure}

We further test the effectiveness of residual perturbation for ResNet8 and En$_5$ResNet8 on the MNIST/CIFAR10/CIFAR100 dataset. Also, we will contrast residual perturbation with DPSGD in privacy protection and utility maintenance.  Fig.~\ref{fig:Cifar10-WithSkip_cn} plots the performance of En$_5$ResNet8 on CIFAR10 under the above four different measures, namely, precision, recall, training and test accuracy, and AUC. Again, the ensemble of ResNets with residual perturbation is remarkably less vulnerable to membership inference attack; e.g., the AUC of the attack model for ResNet8 (non-private) and En$_5$ResNet8 ($\gamma=0.75$) is $0.757$ and $0.573$, respectively. Also, the classification accuracy of En$_5$ResNet8 ($67.5$\%) is higher than that of the non-private ResNet8 ($66.9$\%) for CIFAR10 classification. Fig.~\ref{fig:Cifar100_cn} depicts the results of En$_5$ResNet8 for CIFAR100 classification. These results confirm again that residual perturbation can protect membership privacy and improve classification accuracy.

\paragraph{{Residual perturbation vs. DPSGD}}
{
We contrast residual perturbation with DPSGD on MNIST, CIFAR10, and CIFAR100 classification; the results are shown in Tables~\ref{tab:RP:vs:DPSGD_cn MNIST}, \ref{tab:RP:vs:DPSGD_cn CIFAR10}, and \ref{tab:RP:vs:DPSGD_cn CIFAR100}, respectively. These results unanimously show that 1) both DPSGD and residual perturbation can effectively protect membership privacy, and 2) residual perturbation is better than DPSGD in maintaining the model's utility without sacrificing membership privacy protection.
}

\begin{table}[!ht]
\caption{\small {Residual perturbation {(\textbf{Strategy I})} vs. DPSGD in training ResNet8 and EnResNet8 for MNIST classification. 
Residual perturbation has higher accuracy {without sacrificing membership privacy protection} compared to DPSGD.
}
}\label{tab:RP:vs:DPSGD_cn MNIST}
\centering
\fontsize{9.0pt}{0.9em}\selectfont
\begin{tabular}{c|c|c|c}
\hline
Model  & AUC & Training Acc & Test Acc\\
\hline
{ResNet8(non-private)}   & 0.514 & \bf 1.000 & \bf 0.988\\
ResNet8 (DPSGD) & 0.507 & 0.977 & 0.975\\
{En$_1$ResNet8($\gamma=2.5$)}   & \bf 0.506 & 0.989 & 0.978\\
\hline
\end{tabular}
\end{table}

\begin{table}[!ht]
\caption{\small {Residual perturbation {(\textbf{Strategy I})} vs. DPSGD in training ResNet8 and EnResNet8 for CIFAR10 classification. 
Residual perturbation has higher accuracy 
than the baseline model {without sacrificing membership privacy protection}.
}
}\label{tab:RP:vs:DPSGD_cn CIFAR10}
\centering
\fontsize{9.0pt}{0.9em}\selectfont
\begin{tabular}{c|c|c|c}
\hline
Model  & AUC & Training Acc & Test Acc\\
\hline
{ResNet8(non-private)}   & 0.757 & \bf 1.000 & 0.669\\
ResNet8 (DPSGD) & 0.512 & 0.718 & 0.625\\
{En$_1$ResNet8($\gamma=4.5$)}   & \bf 0.511 & 0.760 & \bf 0.695\\
\hline
\end{tabular}
\end{table}

\begin{table}[!ht]
\caption{\small {Residual perturbation {(\textbf{Strategy I})} vs. DPSGD in training ResNet8 and EnResNet8 for CIFAR100 classification. 
Residual perturbation has higher accuracy than DPSGD and protects comparable membership privacy with DPSGD.
}}\label{tab:RP:vs:DPSGD_cn CIFAR100}
\centering
\fontsize{9.0pt}{0.9em}\selectfont
\begin{tabular}{c|c|c|c}
\hline
Model  & AUC & Training Acc & Test Acc\\
\hline
{ResNet8(non-private)}   & 0.903 & \bf 0.999 & 0.305\\
ResNet8 (DPSGD) & 0.516 & 0.400 & 0.288\\
{En$_1$ResNet8($\gamma=3.25$)}   & \bf 0.515 & 0.485 & \bf 0.383\\
\hline
\end{tabular}
\end{table}

\begin{figure}[t!]
\centering
\begin{tabular}{ccc}
\hskip -0.5cm\includegraphics[clip, trim=0cm 0cm 0cm 0.4cm,width=0.36\columnwidth]{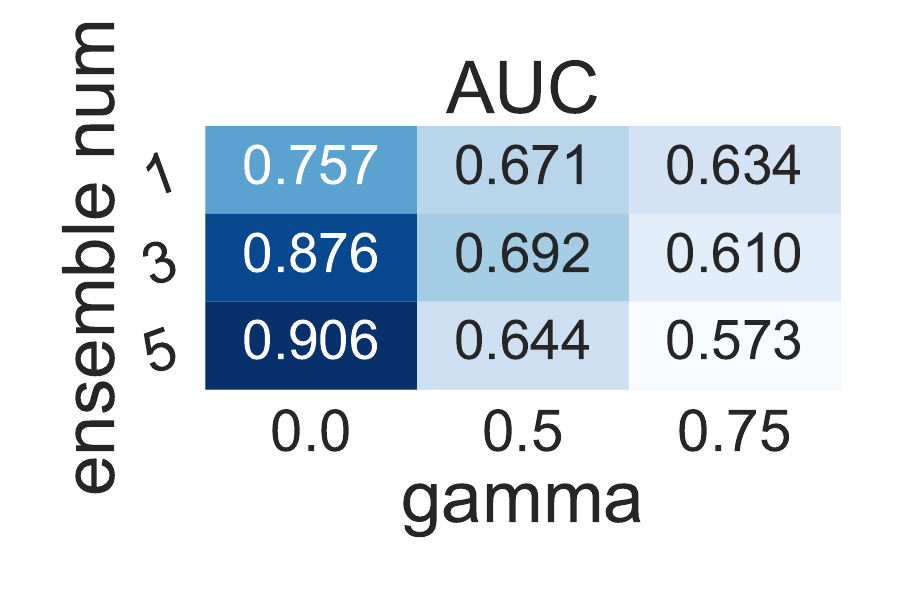}&
\hskip -0.5cm\includegraphics[clip, trim=0cm 0cm 0cm 0.4cm,width=0.36\columnwidth]{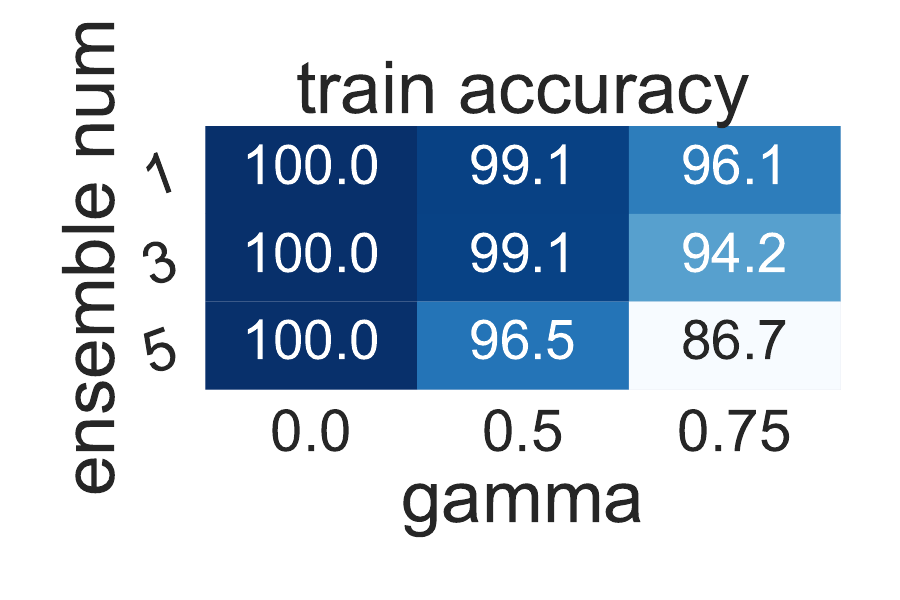}&
\hskip -0.5cm\includegraphics[clip, trim=0cm 0cm 0cm 0.4cm,width=0.36\columnwidth]{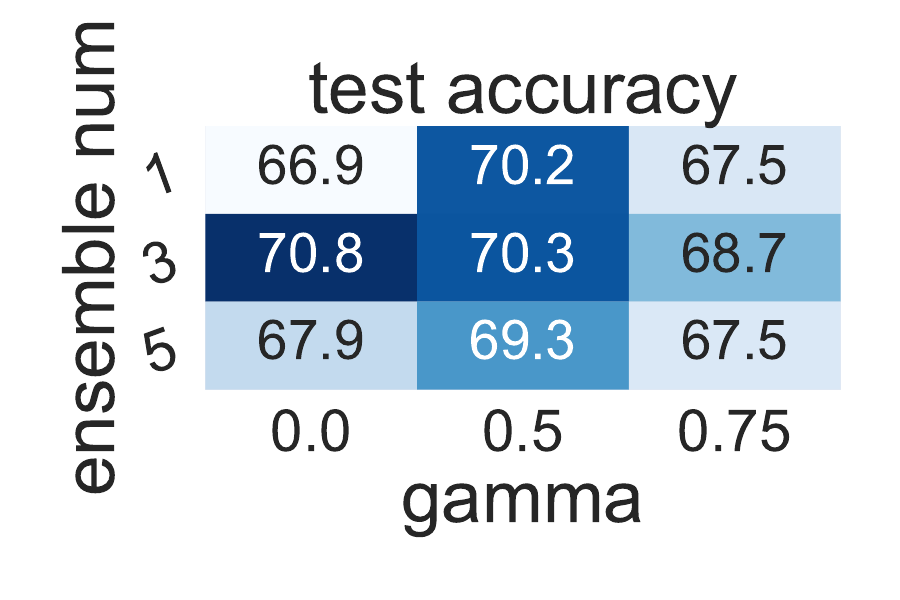}\\
\end{tabular}
\vskip -0.4cm
\caption{\small Performance of residual perturbation {(\textbf{Strategy I})} with different noise coefficients ($\gamma$) and different number of models in the ensemble. 
Privacy-utility tradeoff depends on these two factors.
(Unit: \%)}
\label{fig:ensemble_noise_cn}\vspace{-0.4cm}
\end{figure}

\paragraph{Effects of the number of models in the ensemble}
In this part, we consider the effects of the number of residual perturbed ResNets in the ensemble. Fig.~\ref{fig:ensemble_noise_cn} illustrates the performance of EnResNet8 for CIFAR10 classification measured in the AUC, and training and test accuracy. These results show that tuning the noise coefficient and the number of models in the ensemble is crucial to optimize the tradeoff between accuracy and privacy.

\vspace{-0.3cm}
\subsection{On the Importance of Skip Connections}
Residual perturbations relies on the irreversibility of the SDEs \eqref{Algorithm:Eq4} and \eqref{Algorithm:Eq3}, and this ansatz lies in the skip connections in the ResNet. We test both standard ResNet and the modified ResNet without skip connections. For CIFAR10 classification, under the same noise coefficient ($\gamma=0.75$), the test accuracy is $0.675$ for the En$_5$ResNet8 (with skip connection); while the test accuracy is $0.653$ for the En$_5$ResNet8 (without skip connection). Skip connections makes EnResNet more resistant to noise injection which is crucial for the success of residual perturbation for privacy protection.

\vspace{-0.3cm}
\subsection{ROC Curves for Experiments on Different Datasets}\label{subsection:ROC}
\begin{figure}[t!]
\centering
\begin{tabular}{ccc}
\hskip -0.5cm\includegraphics[clip, trim=0cm 2cm 0cm 0.15cm,width=0.37\columnwidth]{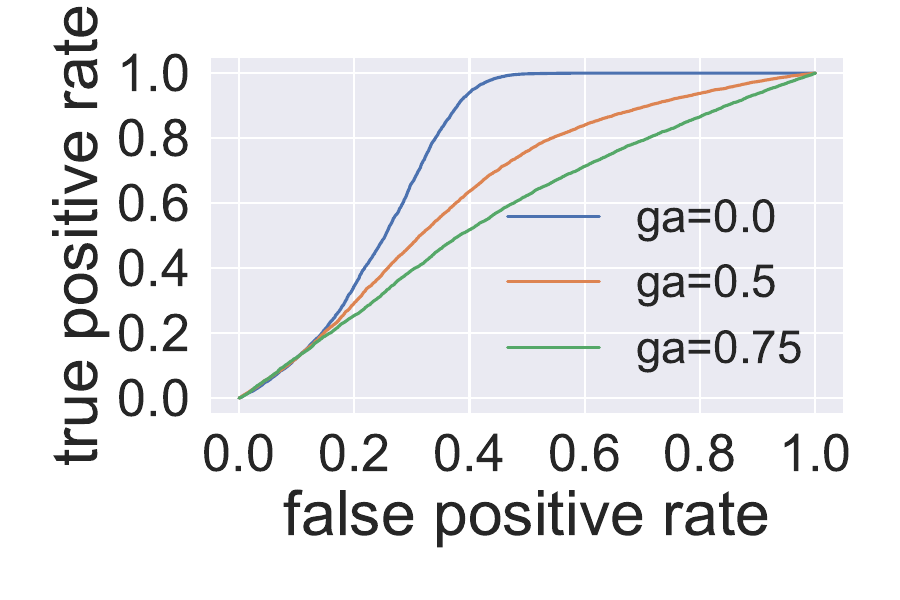}&
\hskip -0.65cm\includegraphics[clip, trim=0cm 2cm 0cm 0.15cm,width=0.37\columnwidth]{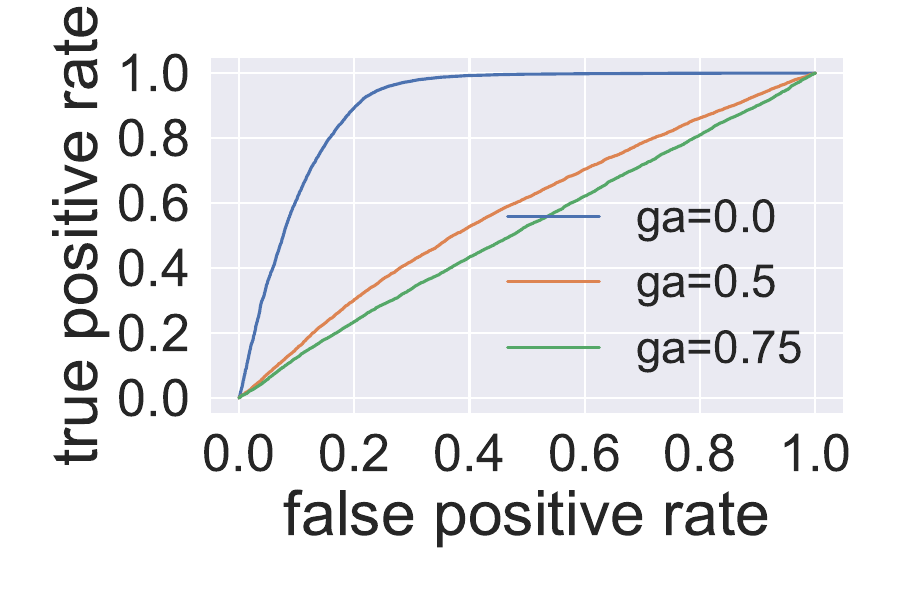}&
\hskip -0.65cm\includegraphics[clip, trim=0cm 2cm 0cm 0.15cm,width=0.37\columnwidth]{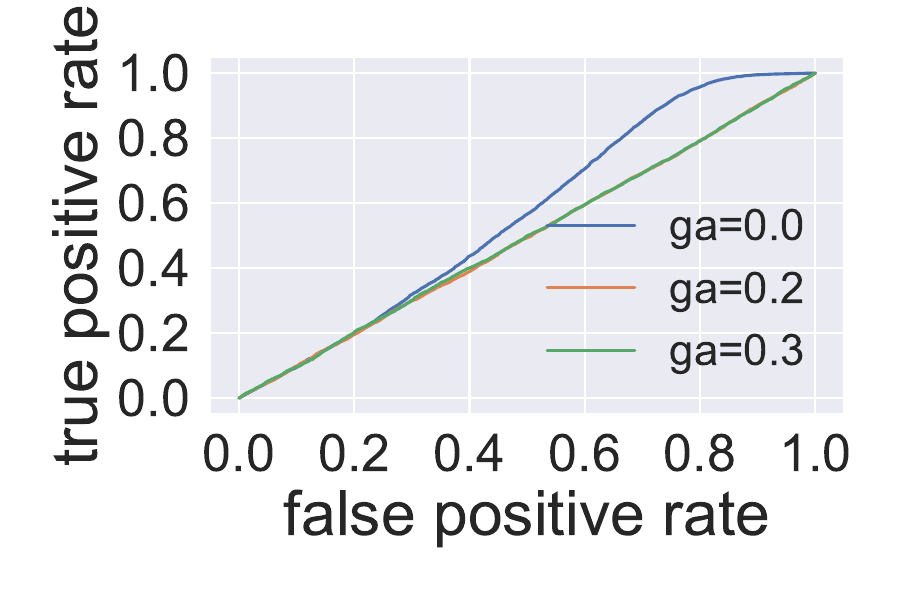}\\
{\small \hskip -0.2cm (a) CIFAR10} & {\small \hskip -0.2cm (b) CIFAR100} & {\small \hskip -0.2cm (c) IDC}\\
{\small \hskip -0.2cm (En$_5$ResNet8)} & {\small \hskip -0.2cm  (En$_5$ResNet8)} & {\small \hskip -0.2cm (En$_5$ResNet8)}\\
\end{tabular}
\vskip -0.2cm
\caption{\small ROC curves for residual perturbation {(\textbf{Strategy I})} for different datasets. (ga: noise coefficient $\gamma$)}
\label{fig:ROC_cn}\vspace{-0.4cm}
\end{figure}

The receiver operating characteristic (ROC) curve can be used to illustrate the classification ability of a binary classifier. ROC curve is obtained by plotting the true positive rate against the false positive rate at different thresholds. The true positive rate, also known as recall, is the fraction of the positive set (all the positive samples) that is correctly inferred as a positive sample by the binary classifier. The false positive rate can be calculated by 1-specificity, where specificity is the fraction of the negative set (all the negative samples) that is correctly inferred as a negative sample by the binary classifier. In our case, the attack model is a binary classifier. Data points in the training set of the target model are tagged as positive samples, and data points out of the training set of the target model are tagged as negative samples. Then we plot ROC curves for different datasets (as shown in Fig.~\ref{fig:ROC_cn}). These ROC curves show that if $\gamma$ is sufficiently large, the attack model's prediction will be nearly a random guess.

\vspace{-0.3cm}
\subsection{Remark on the Privacy Budget}
{In the experiments above, we set the constants $G$ and $R$ to $30$ for {\bf Strategy I}. For classifying IDC with ResNet8, {the theoretical DP budget, based on Theorem~\ref{theorem1}}, for {\bf Strategy I} is $\small (\epsilon=1.1e3,\delta=1e-5)$ and the DP-budget for DPSGD is $\small (\epsilon=15.79,\delta=1e-5)$. For classifying CIFAR10 with ResNet8, the DP budget, based on Theorem~\ref{theorem1}, for {\bf Strategy I} is $(\epsilon=3e3,\delta=1e-5)$ and the DP-budget for DPSGD is $\small (\epsilon=22.33,\delta=1e-5)$.} {Using the algorithm provided in \cite{jagielski2020auditing}, we estimate the empirical lower bound of the $\epsilon$ given $\delta=1e-5$
for both {\bf Strategy I} and DPSGD. The obtained empirical lower bound can guide the improvement of analytical tools to bound the DP budget --- the larger gap between theoretical and empirical DP budget indicates more theoretical budget to improve. For IDC classification, the lower bound of $\epsilon$ for {\bf Strategy I} and DPSGD are {3.604e-3} and {6.608e-3}, respectively. And the lower bound of $\epsilon$ for {\bf Strategy I} and DPSGD for CIFAR10 classification are {1.055e-1} and {1.558e-1}, respectively.} 
{These empirical DP budget estimates show that Theorem~\ref{theorem1} offers a quite loose DP budget compared to DPSGD.} There are several challenges we need to overcome to get a tighter DP bound for {\bf Strategy I}. Compared to DPSGD, it is significantly harder. In particular, 1) the loss function of the noise injected ResNets is highly nonlinear and very complex with respect to the weights, also the noise term appears in the loss function due to the noise injected in each residual mapping. 
2) In our proof, we leverage the framework of subsampled R\'enyi-DP \cite{wang2018subsampled} to find a feasible range of noise variance parameter, and then convert to DP to get the value of $\gamma$ for a given DP budget. This procedure will significantly reduce the accuracy of the estimated $\gamma$. We leave the tight DP guarantee as future work. In particular, how to reduce the accuracy of estimating  due to the conversion between R\'enyi-DP and DP.  

\vspace{-0.3cm}
\section{Technical Proofs}\label{sec:proofs}
\subsection{Proof of Theorem~\ref{theorem1}}\label{sec:appendix:Thm1}
\subsubsection{R\'enyi Differential Privacy (RDP)}
We use the notion of RDP to prove the DP guarantees for residual perturbations.
\begin{definition}\cite{mironov2017renyi}
	(R\'{e}nyi divergence) For any two probability distributions $P$ and $Q$ defined over the distribution $\mathcal D$, the R\'{e}nyi divergence of order $\alpha>1$ is
	\begin{gather*}\small
D_{\alpha}(P||Q)=\frac{1}{\alpha-1}\log\EE_{x\sim Q}(P/Q)^{\alpha}.
	\end{gather*}
\end{definition}
\begin{definition}\cite{mironov2017renyi}
($(\alpha,\epsilon)$-RDP) A randomized mechanism $\mathcal M:\mathcal D\rightarrow\mathcal R$ is $\epsilon$-R\'{e}nyi DP of order $\alpha$ or $(\alpha,\epsilon)$-RDP, if for any adjacent $\mathcal S,\mathcal S^{'}\in\mathcal D$ that differ by one entry, we have 
\begin{gather*}\small
D_{\alpha}(\mathcal M(\mathcal S)||\mathcal M(\mathcal S^{'})\leq \varepsilon.
	\end{gather*}
\end{definition}

\begin{lemma}\label{appendix:lemma1}\cite{mironov2017renyi}
Let $f:\mathcal D\rightarrow\mathcal R_{1}$ be $(\alpha,\epsilon_{1})$-RDP and $g:\mathcal R_{1}\times\mathcal D\rightarrow\mathcal R$ be $(\alpha,\epsilon)$-RDP,  then the mechanism $(X,Y)$ with $X\sim f(D)$ and $Y\sim g(X,D)$, satisfies $(\alpha,\epsilon_{1}+\epsilon_{2})$-RDP.
\end{lemma}
\begin{lemma}\label{appendix:lemma2}\cite{mironov2017renyi}
(From RDP to $(\epsilon,\delta)$-DP) If $f$ is an $(\alpha,\epsilon)$-RDP mechanism, then it also satisfies $(\epsilon-(\log\delta)/(\alpha-1),\delta)$-differential privacy
for any $0<\delta<1$.
\end{lemma}
\begin{lemma}\label{appendix:lemma3}\cite{mironov2017renyi}
(Post-processing lemma) Let $\mathcal M:\mathcal D\rightarrow\mathcal R$ be a randomized algorithm that is $(\alpha,\epsilon)$-RDP, and let $f:\mathcal R\rightarrow\mathcal R^{'}$ be an arbitrary randomized mapping. Then $f(\mathcal M(\cdot)):\mathcal D\rightarrow \mathcal R^{'}$ is $(\alpha,\epsilon)$-RDP.
\end{lemma}

\vspace{-0.1cm}
\subsubsection{Proof of Theorem~\ref{theorem1}}
In this subsection, we will give a proof of Theorem~\ref{theorem1}, i.e., DP-guarantee for the {\bf Strategy I}.
\vspace{-0.3cm}
\begin{proof}
We prove Theorem~\ref{theorem1} by mathematical induction. Consider two adjacent datasets $\mathcal S=\{\vx_1,\cdots,\vx_{N-1},\vx_N\},\mathcal S'=\{\vx_1,\cdots,\vx_{N-1},\hat\vx_N\}$ that differ by one entry. 
For the first residual mapping, it is easy to check that when {$\pi\geq R\sqrt{2\alpha/\epsilon_p}$} {with $\pi$ being the standard deviation of the Gaussian noise injected into the input data}, we have {$D_{\alpha}(\vx^0_N||\hat\vx^0_N)=\alpha\|\vx_N-\hat\vx_N\|^2/(2\pi^2)<\varepsilon_p$}. For the remaining residual mappings, we denote the response of the $i$-th residual mapping, for any two input data $\vx_N$ and $\hat\vx_N$, as $\vx^i_N,\hat\vx^{i}_N$, respectively. Based on our assumpution, we have 
\begin{equation}\small\begin{aligned}
\vx^i_N+\phi(\mU^i\vx^i_N)&\sim\mathcal N(\mu_{N,i},\sigma^2_{N,i}),\\ 
\hat\vx^i_N+\phi(\mU^i\hat\vx^i_N)&\sim\mathcal N(\hat\mu_{N,i},\sigma^2_{N,i}),
\end{aligned}
\end{equation}
where {the mean} $\mu_{N,i}$ and $\hat\mu_{N,i}$ are both bounded by the constant $G$. If {$D_{\alpha}(\vx^i_N||\hat\vx^i_N)\leq \epsilon_p/(i+1)$}, according the post-processing lemma (Lemma~\ref{appendix:lemma3}), we have 
{\begin{equation}\label{eq:app1}\small
\begin{aligned}
&D_{\alpha}(\vx^i_N+\phi(\mU^i\vx^i_N)||\hat\vx^i_N+\phi(\mU^i\hat\vx^i_N))\\
=&\frac{\alpha\|\mu_{N,i}-\mu^{'}_{N,i}\|^2}{2\sigma^2_{N,i}}
\leq \epsilon_p/(i+1),
\end{aligned}
\end{equation}}
so when {the standard deviation of the injected Gaussian noise satisfies} $\gamma>\sqrt{2\alpha G^2/\epsilon_p}$, \eqref{eq:app1} implies that
{\begin{equation}\small
\begin{aligned}
&D_{\alpha}(\vx^i_N+\phi(\mU^i\vx^i_N)+\gamma\vn||\hat\vx^i_N+\phi(\mU^i\hat\vx^i_N)+\gamma\vn)\\
=&\frac{\alpha\|\mu_{N,i}-\mu^{'}_{N,i}\|^2}{2(\sigma^2_{N,i}+\gamma^2)}\leq \frac{\epsilon_p}{i+2},
\end{aligned}
\end{equation}}
implying that $D_{\alpha}(\vx^{i+1}_{N}||\hat\vx^{i+1}_{N})<\frac{\epsilon_p}{i+2}$. 
Moreover, note that
\begin{equation}\small
\begin{aligned}
\hspace{-0.5cm}\nabla_{\mU^{i+1}}\ell|_{\vx^{i+1}_{N}}&=l^{'}\left(\vx^{i+1}_{N}+\phi\left(\mU^{i+1}\vx^{i+1}_{N}\right)\right)\phi^{'}(\mU^{i+1}\vx^{i+1}_{N})(\vx^{i+1}_{N})^{\top}\\
\nabla_{\mU^{i+1}}\ell|_{\hat\vx^{i+1}_{N}}&=l^{'}\left(\hat\vx^{i+1}_{N}+\phi\left(\mU^{i+1}\hat\vx^{i+1}_{N}\right)\right)\phi^{'}(\mU^{i+1}\hat\vx^{i+1}_{N})(\hat\vx^{i+1}_{N})^{\top}.
\end{aligned}
\end{equation}
By the post-processing lemma (Lemma~\ref{appendix:lemma3}) again, we get
{\begin{equation*}\small
D_{\alpha}(\nabla_{\mU^{i+1}}\ell|_{\vx^{i+1}_{N}}||\nabla_{\mU^{i+1}}\ell|_{\hat\vx^{i+1}_{N}})<\frac{\epsilon_p}{i+2}.
\end{equation*}}
Let $\mathcal B_t$ be the index set with $|\mathcal B_t|=b$, and we update $\mU^i$ as 
\begin{equation}\small
\begin{aligned}
\mU^{i+1}_{t+1}|\mathcal S&=\mU^{i}_{t}-\alpha \frac{1}{b}\sum_{j\in\mathcal B_t}\nabla_{\mU^{i+1}_t}\ell(\vx^{i+1}_{j},\mU^{i+1}_t),\quad
\\
\mU^{i+1}_{t+1}|\mathcal S'&=\mU^{i}_{t}-\alpha \frac{1}{b}\sum_{j\in\mathcal B_t}\nabla_{\mU^{i+1}_t}\ell(\vx^{i+1}_{j},\mU^{i+1}_t),
\end{aligned}
\end{equation}
where $\mU^{i+1}_t$ is the weights updated after the $t$-th training iterations. When $N\notin\mathcal B_t$, it's obviously that $D_{\alpha}(\mU^{i+1}_{t+1}|\mathcal S||\mU^{i+1}_{t+1}|\mathcal S')=0$; when $N\in\mathcal B_t$, the equations which we use to update $\mU^i$ can be rewritten as
\begin{equation}\small
\begin{aligned}
\mU^{i+1}_{t+1}|\mathcal S&=\mU^{i}_{t}-\alpha (1/b)\sum_{j\in\mathcal B_t-\{N\}}\nabla_{\mU^{i+1}_t}\ell(\vx^{i+1}_{j},\mU^{i+1}_t)\\
&\ \ -\alpha(1/b)\nabla_{\mU^{i+1}_t}\ell(\vx^{i+1}_{N},\mU^{i+1}_t)\\
\mU^{i+1}_{t+1}|\mathcal S'&=\mU^{i}_{t}-\alpha (1/b)\sum_{j\in\mathcal B_t-\{N\}}\nabla_{\mU^{i+1}_t}\ell(\vx^{i+1}_{j},\mU^{i+1}_t)\\
&\ \ -\alpha(1/b)\nabla_{\mU^{i+1}_t}\ell(\hat\vx^{i+1}_{N},\mU^{i+1}_t).
\end{aligned}
\end{equation}
By Lemma~\ref{appendix:lemma3},
we have {$D_{\alpha}(\mU^{i+1}_{t+1}|\mathcal S||\mU^{i+1}_{t+1}|\mathcal S')\leq\epsilon_p/(i+2)$}. Because there are only $(Tb)/N$ steps where we use the information of $\vx_N$ and $\hat\vx_N$. Replacing $\epsilon_p$ by $(N\epsilon_p)/(Tb)$ and use composition theorem we can get after $T$ steps the output $\mU^{i+1}_T$ satisfies {$(\alpha,\epsilon_p/(i+2))$}-RDP and $\vw$ satisfies {$(\alpha,\epsilon_p/(M+1))$}-RDP. By Lemma~\ref{appendix:lemma2}, we can easily establish the DP-guarantee for {\bf Strategy I}, as stated in Theorem~\ref{theorem1}.
\end{proof}

\vspace{-0.3cm}
\subsection{Proof of Theorem~\ref{thm:dp}}\label{sec:appendix:Thm2}
In this section, we will provide a proof for Theorem~\ref{thm:dp}.

\vspace{-0.1cm}
\begin{proof}
{Let $\phi=BN(\psi)$, where BN is the batch normalization operation and $\psi$ is an activation function. 
Due to the property of batch normalization, we assume {the $\ell_2$ norm of} $\phi$ can be bounded by a positive constant $B$}. To show that the model \eqref{EnResNet1} guarantees training data privacy given only black-box access to the model. Consider training the model \eqref{EnResNet1} with two different datasets $S$ and $S'$, and we denote the 
resulting model as $f(\cdot|S)$ and $f(\cdot|S')$, respectively. 
Next, we show that using appropriate 
$\gamma$ and $\pi$, $D_{\alpha}(f(\vx|S)||f(\vx|S^{'}))<\epsilon_{p}$ for any input $\vx$. 

First, we consider the convolution layers, and let ${\rm Conv}(\vx)_{i}$ be the $i$-th entry of the vectorized ${\rm Conv}(\vx)$. Then we have {${\rm Conv}(\vx)_{i}=\vx_{i}+\phi(\mU\vx)_{i}+\gamma \tilde{\vx}_{i}\odot\vn_{i}$}. 
{For any two different training datasets, we denote
\begin{equation*}\small
\begin{aligned}
{\rm Conv}(\vx)_{i}| S&=\vx_{i}+\phi(\mU_{1}\vx)_{i}+\gamma \tilde{\vx}_{i}\odot\vn_{i}\\
&\sim\mathcal N(\vx_{i}+\phi(\mU_{1}\vx)_{i},\gamma^{2} \bX),
\end{aligned}
\end{equation*}
and 
\begin{equation*}\small
\begin{aligned}
{\rm Conv}(\vx)_{i}| S^{'}&= \vx_{i}+\phi(\mU_{2}\vx)_{i}+\gamma \tilde{\vx}_{i}\odot\vn_{i}\\
&\sim\mathcal N(\vx_{i}+\phi(\mU_{2}\vx)_{i},\gamma^{2} \bX),
\end{aligned}
\end{equation*}}
{where $\bX$ is a diagonal matrix with the $j$-th diagonal entry 
being the square of the $j$-th element of the vector $\Tilde{\bm x}^{2}_{i}$.}

Therefore, if $\gamma>(B/\eta)\sqrt{(2\alpha M)/(\epsilon_{p})}$, we have
\begin{equation*}\small
\begin{aligned}
 & D_{\alpha}\left(\mathcal N\left(\vx_{i}+\phi\left(\mU_{1}\vx\right)_{i},\gamma^{2} \tilde{\vx}^{2}_{i}\right)||\mathcal N\left(\vx_{i}+\phi\left(\mU_{2}\vx\right)_{i},\gamma^{2} \tilde{\vx}^{2}_{i}\right)\right)\\
 &\leq\frac{\alpha(\phi(\mU_{1}\vx)_{i}-\phi(\mU_{2}\vx)_{i})^{2}}{2\gamma^{2}\eta^2}\\
 &\leq \frac{4\alpha L^{2}B^{2}\|\vx\|^2_{2}}{2\gamma^{2}\eta^{2}}
 \leq \frac{2\alpha B^{2}}{\gamma^{2}\eta^{2}}\leq \epsilon_{p}/M.
\end{aligned}
 \end{equation*}
Furthermore, $\gamma>(b/\eta)\sqrt{(2\alpha M)/(\epsilon_{p})}$ guarantees $(\alpha,\epsilon_{p}/M)$-RDP for every convolution layer. For the last fully connected layer, if $\pi>a\sqrt{\left(2\alpha M\right)/\epsilon_{p}}$, we have
\begin{equation*}\small
\begin{aligned}
 & \ \ \ \ D_{\alpha}\left(\mathcal N\left(\vw_{1}^{\top}\vx,\pi^{2} \tilde{\vx}^{2}_{i}\right)||\mathcal N\left(\vw_{2}^{\top}\vx,\pi^{2} \tilde{\vx}^{2}_{i}\right)\right)\\
 &=\frac{\alpha\left(\vw_{1}^{\top}\vx-\vw_{2}^{\top}\vx\right)^{2}}{2\pi^{2}\|\vx\|_{2}}\leq \frac{4\alpha a^{2}\|\vx\|^2_{2}}{2\pi^{2}\|\vx\|_{2}}
 \leq \frac{2\alpha a^{2}}{\pi^{2}}\leq \epsilon_{p}/M,
\end{aligned}
\end{equation*}
i.e., $\pi>a\sqrt{\left(2\alpha M\right)/\epsilon_{p}}$ guarantees that the fully connected layer to be $(\alpha,\epsilon_{p}/M)$-RDP. According to Lemma~\ref{appendix:lemma1}, we have $(\alpha,\epsilon_{p})$-RDP guarantee for the ResNet of $M$ residual mappings if $\gamma>(Lb/\eta)\sqrt{(2\alpha dM)/(\epsilon_{p})}$ and $\pi>a\sqrt{\left(2\alpha M\right)/\epsilon_{p}}$. Let $\lambda\in(0,1)$, for any given  $(\epsilon_{p}, \delta)$ pair, if 
$\epsilon_{p}\leq\lambda\epsilon$ and $-(\log\delta)/(\alpha-1),\delta)\leq(1-\lambda)\epsilon$, then we have get the $(\epsilon, \delta)$-DP guarantee for the ResNet with $M$ residual mapping using the residual perturbation {\bf Strategy II}.
\end{proof}

\vspace{-0.3cm}
\subsection{Proof of Theorem~\ref{Theory:Thm1}}\label{sec:appendix:Thm3}
In this section, we will proof that the residual perturbation~\eqref{eq:ResidualPerturbation2} can reduce the generalization error via computing the Rademacher complexity. Let us first recap on some related lemmas on SDE and Rademacher complexity.

\subsubsection{Some Lemmas}
Let $\ell:V\times Y\rightarrow[0,B]$ be the loss function. Here we assume $\ell$ is bounded and $B$ is a positive constant. In addition, we denote the function class $\ell_{\mathcal H}=\{(\vx,y)\rightarrow \ell(h(\vx),y):h\in\mathcal H, (\vx,y)\in X\times Y\}$. The goal of the learning problem is to find $h\in\mathcal H$ such that the population risk $R(h)=\mathbb E_{(\vx,y)\in\mathcal D}[\ell(h(\vx),y)]$ is minimized. The gap between population risk and empirical risk $R_{S_{N}}(h)=(1/N)\sum^{N}_{i=1}\ell(h(\vx_{i}),y_{i})$ is known as the generalization error. We have the following lemma and theorem to connect the population and empirical risks via Rademacher complexity.

\begin{lemma}
(Ledoux-Talagrand inequality)
\cite{ledoux2002probability}Let $\mathcal H$ be a bounded real valued function space and let $\phi: \mathbb R\rightarrow \mathbb R$ be a Lipschitz with constant L and $\phi (0)=0$. Then we have
\begin{equation*}\small
\frac{1}{n}E_{\sigma}\left[\sup_{h\in \mathcal H}\sum_{i=1}^n\sigma_i\phi \left(h\left(x_i\right)\right)\right]\leq \frac{L}{n}E_{\sigma}\left[\sup_{h\in \mathcal H}\sum_{i=1}^n\sigma_i h\left(x_i\right)\right].
\end{equation*}
\end{lemma}

\begin{lemma}
\cite{peter2002rademacher}Let $S_{N} = \{\left(\vx_1, y_1\right),\cdots, \left(\vx_N, y_N\right)\}$ be samples chosen i.i.d. according to the distribution $\mathcal{D}$. If the loss function $\ell$ is bounded by $B>0$. Then for any $\delta\in (0, 1)$, with probability at least $1-\delta$, the following holds for all $h\in \mathcal H$,
\begin{equation*}\small
R\left(h\right) \leq R_{S_{N}}\left(h\right) + 2BR_{S_{N}}\left(\ell_{\mathcal H}\right)+3B\sqrt{(\log (2/\delta)/(2N)}.
\end{equation*}
In addition, according to the Ledoux-Talagrand inequality and assume loss function is $L$-lipschitz, we have
\begin{equation*}\small
R_{S_{N}}\left(\ell_{\mathcal H}\right)\leq LR_{S_{N}}\left(\mathcal H\right).
\end{equation*}
\end{lemma}

So the population risk can be bounded by the empirical risk and Rademacher complexity of the function class $\mathcal H$. Because we can't minimize the population risk directly, we can minimize it indirectly by minimizing the empirical risk and Rademacher complexity of the function class $\mathcal H$. Next, we will further discuss Rademacher complexity of the function class $\mathcal H$. We first introduce several lemmas below.

\begin{lemma}\label{C5}\small
\cite{fima2002introduction} For any given matrix $\mU$, the solution to the equation 
\begin{equation*}\small
d\vx(t)=\mU\vx(t)dt;\ 
			\vx(0) = \hat{\vx},
\end{equation*}
has the following expression
\begin{equation*}\small
\vx(t) = \exp (\mU t)\hat{\vx},
\end{equation*}
Also, we can write the solution to the following equation
\begin{equation*}\small
d\vy(t)=\mU\vy(t)dt + \gamma \vy(t)dB(t);\ \vy(0) = \hat{\vx},
\end{equation*}
as
$$\small
\vy(t) = \exp (\mU t-\frac{1}{2}\gamma^2 t \mI + \gamma B(t)\mI)\hat{\vx}.
$$
Obviously, we have $\EE[\vy(t)]=\vx(t)$.
\end{lemma}

\begin{lemma}\label{C7}
For any circulant matrix $\Cb$ with the first row being $(a_1,a_2,\cdots,a_d)$, we have the following eigen-decomposition
$\Psi ^H \Cb\Psi=diag(\lambda_1,\dots,\lambda_d),$
where
\begin{equation*}\small
	\sqrt{d}\Psi = \begin{pmatrix}
		1&1&\dots&1\\
		1&m_1&\dots&m_{d-1}\\
		\dots&\dots&\dots&\dots\\
		1&m^{d-1}_1&\dots&m^{d-1}_{d-1}
	\end{pmatrix},
\end{equation*}
and $m_i$s are the roots of unity and $\lambda_i=a_1+a_2 m_i+\dots+a_d m^{d-1}_i$.
\end{lemma}

\vspace{-0.3cm}
\subsubsection{The proof of Theorem~\ref{Theory:Thm1}}
\vspace{-0.2cm}
\begin{proof}
By the definition of Rademacher complexity (Def.~\ref{def:Rademacher:Complexity}), we have
\begin{equation*}\small
\begin{aligned}
R_{S_N}\left(\mathcal F\right) &= \left(1/N\right)\mathbb E_{\sigma}\left[\sup_{f\in \mathcal F}\sum_{i=1}^N\sigma_i\vw\vx^p_i(T)\right]\\
&= \left(c/N\right)\mathbb E_{\sigma}\left[\sup_{\|U\|_2\leq c}\|\sum_{i=1}^N\sigma_i \vx^p_i\left(T\right)\|_2\right].
\end{aligned}
\end{equation*}
Let $\vu_i=\Psi \vx^p_i$ and denote the $j$-th element of $\vu_i$ as $u_{i,j}$. Then by lemma \ref{C7}, we have
\begin{equation*}\small
\begin{aligned}
		&R_{S_N}\left(\mathcal F\right)\\
		&=\left(c/N\right)\mathbb E_{\sigma}\sup_{\|\mU\|_2\leq c}\left(\sum_{i, j}\sigma_i\sigma_j\langle \vx^p_i(T), \vx^p_j(T)\rangle \right)^{1/2}\\
	    & =\left(c/N\right)\mathbb E_{\sigma}\sup_{|\lambda_i|\leq c}(\|\Psi^H \sum_{i=1}^{N}\sigma_i\big(u_{i,1}\exp (\lambda_1 Tp),\\
	    &\ \ \ \ \ \ \ \ \ \ \ \ \ \ \ \ \cdots, u_{i, d}\exp (\lambda_d Tp))^\top\|_2\big)\\
		& = \left(c/N\right)\mathbb E_{\sigma}\sup_{|\lambda_i|\leq c}\{\sum_{j=1}^d\left[\sum_{i=1}^N\sigma_i \vu_{i, j}\exp \left(\lambda_j Tp\right)\right]^2\}^{1/2}\\ 
		&= \left(c/N\right)\exp \left(cTp\right)\mathbb E_{\sigma}\|\sum_{i=1}^N\sigma_i\vu_{i}\|_2\\
		&= \left(c/N\right)\exp \left(cTp\right)\mathbb E_{\sigma}\|\Psi\sum_{i=1}^N\sigma_i\vx^p_{i}\|_2\\
		&= (c/N)\exp \left(cTp\right)\mathbb E_{\sigma}\|\sum_{i=1}^N\sigma_i\vx^p_{i}\|_2
\end{aligned}
\end{equation*}
Note that $\mathbb E\left(\vw\vx^p_i(T)\right)=\vw\mathbb  E\left(\vx^p_i(T)\right)$ and according to Lemma \ref{C5}, similar to proof for the function class $\mathcal F$ we have
\begin{equation*}\footnotesize
\begin{aligned}
\hspace{-0.5cm}&R_{S_N}\left(\mathcal G\right)\\
=&\left(c/N\right)\mathbb E_{\sigma}\sup_{\|\mU\|_2\leq c}\left(\sum_{i, j}\sigma_i\sigma_j\langle \mathbb E\vx^p_i(T), \mathbb E\vx^p_j(T)\rangle \right)^{1/2}\\
=&(c/N)\mathbb E_{\sigma}\sup_{|\lambda_i|\leq c}(\|\Psi^H \sum_{i=1}^{N}\sigma_i(u_{i,1}\exp (\lambda_1 Tp-p(1-p)\gamma^2T/2),\\ &\hskip 2.5cm \cdots, u_{i, d}\exp (\lambda_d Tp-p(1-p)\gamma^2T/2))^\top\|_2)\\
=&\left(c/N\right)\mathbb E_{\sigma}\sup_{|\lambda_i|\leq c}\Big\{\sum_{j=1}^d\Big[\sum_{i=1}^N\sigma_i \vu_{i, j}\exp\times \\
&\hskip 2.5cm \big(\lambda_j Tp-p(1-p)\gamma^2T/2\big)\Big]^2\Big\}^{1/2} \\
=&(c/N)\exp (cTp-p(1-p)\gamma^2T/2)\mathbb E_{\sigma}\Big\{\sum_{j=1}^d\left[\sum_{i=1}^N\sigma_iu_{i, j}\right]^2\Big\}^{1/2}\\
=&\left(c/N\right)\exp \left(cTp-p(1-p)\gamma^2T/2\right)\mathbb E_{\sigma}\|\sum_{i=1}^N\sigma_i\vu_{i}\|_2\\
=&\left(c/N\right)\exp \left(cTp-p(1-p)\gamma^2T/2\right)\mathbb E_{\sigma}\|\Psi\sum_{i=1}^N\sigma_i\vx^p_{i}\|_2\\
=&(c/N)\exp \left(cTp-p(1-p)\gamma^2T/2\right)\mathbb E_{\sigma}\|\sum_{i=1}^N\sigma_i\vx^p_{i}\|_2\\
<&R_{S_N}\left(\mathcal F\right)
\end{aligned}
\end{equation*}
Therefore, we have completed the proof for the fact that the ensemble of Gaussian noise injected ResNets can reduce generalization error compared to the standard ResNet. 
\end{proof}

\vspace{-0.3cm}
\section{Experiments on Strategy II}\label{sec:appendix:strategyII}
\subsection{Forward and Backward Propagation Using \eqref{Algorithm:Eq3}}
\label{sec:appendix:multiplicative:noise}
\begin{figure}[t!]
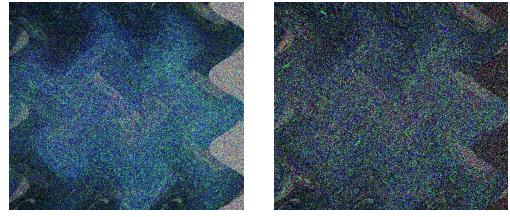

\centering
\begin{tabular}{cc}
\includegraphics[clip, trim=0cm 1cm 0cm 0cm,width=0.35\columnwidth]{sde1.jpg}&
\includegraphics[clip, trim=0cm 1cm 0cm 0cm,width=0.35\columnwidth]{sde2.jpg}\\
{(a) $\vx(1)$ (SDE)}  & {(b) $\Tilde{\vx}(0)$ (SDE)}\\
\end{tabular}
\vskip -0.2cm
\caption{\small Illustrations of the forward and backward propagation of the training data using SDE model \eqref{Algorithm:Eq3}. (a) is the features of the original image generated by the forward propagation using SDE; (b) is the recovered images by reverse-engineering the features shown in (a).}
\label{fig:ODE:vs:SDE-Privacy-sde2}\vspace{-0.4cm}
\end{figure}

Fig.~\ref{fig:ODE:vs:SDE-Privacy-sde2} plots the forward and backward propagation of the image using the SDE model~\eqref{Algorithm:Eq3}. Again, we cannot simply use the backward Euler-Maruyama discretization to reverse the features generated by propagating through the forward Euler-Maruyama discretization.

\subsubsection{Experiments on the IDC Dataset}
\begin{figure}[!ht]
\centering
\begin{tabular}{cc}
\hskip -0.45cm\includegraphics[clip, trim=0cm 0cm 0cm 0.8cm,width=0.48\columnwidth]{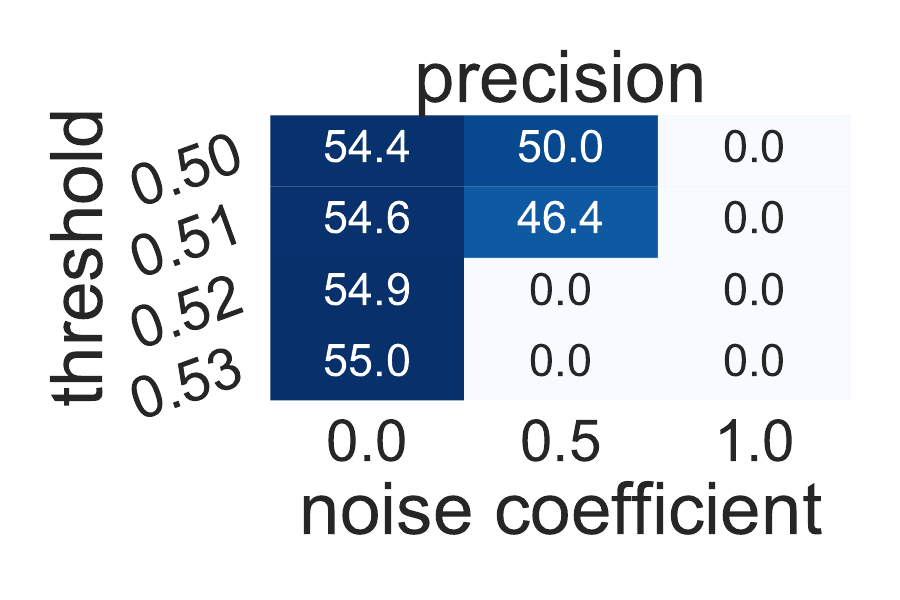}&
\hskip -0.45cm\includegraphics[clip, trim=0cm 0cm 0cm 0.8cm,width=0.48\columnwidth]{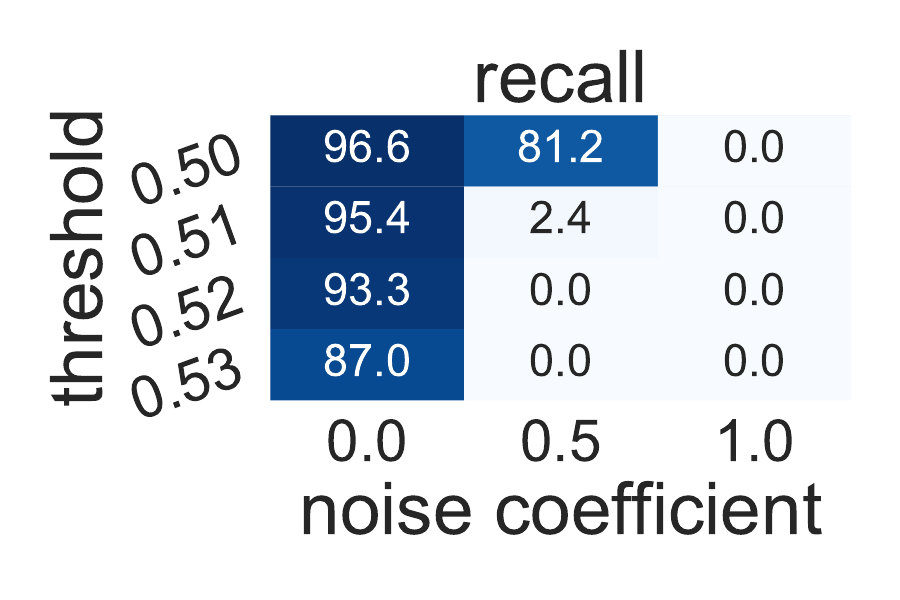}\\
\hskip -0.45cm\includegraphics[clip, trim=0cm 0cm 0cm 0.8cm,width=0.48\columnwidth]{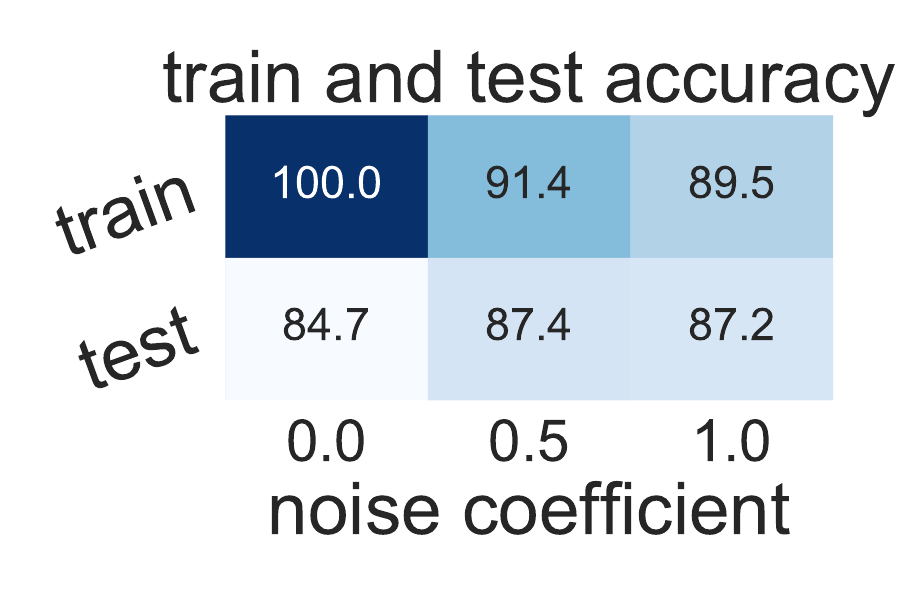}&
\hskip -0.45cm\includegraphics[clip, trim=0cm 0cm 0cm 0.8cm,width=0.48\columnwidth]{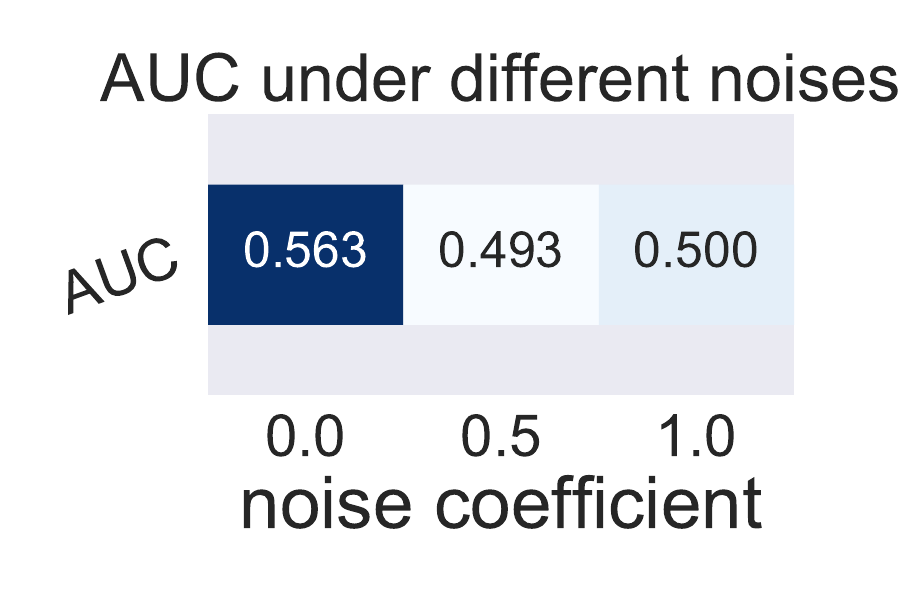}\\
\end{tabular}
\vskip -0.3cm
\caption{\small Performance of residual perturbation~\eqref{eq:ResidualPerturbation2} {(\textbf{Strategy II})} for En$_5$ResNet8 with different noise coefficients ($\gamma$) and membership inference attack thresholds on the IDC dataset. Residual perturbation can significantly improve membership privacy and reduce the generalization gap. $\gamma=0$ corresponding to the baseline ResNet8. (Unit: \%)}
\label{fig:IDC}\vspace{-0.4cm}
\end{figure}

In this subsection, we consider the performance of residual perturbation {(\textbf{Strategy II})} in protecting membership privacy while retaining the classification accuracy on the IDC dataset. We use the same ResNet models as that used for the first residual perturbation. We list the results in Fig.~\ref{fig:IDC}, these results confirm that the residual perturbation~\eqref{eq:ResidualPerturbation2} can effectively protect data privacy and maintain or even improve the classification accuracy. In addition, we depict the ROC curve for this experiment in Fig.~\ref{fig:ROC}(c). We note that, as the noise coefficient increases, the gap between training and testing accuracies narrows, which is consistent with Theorem~\ref{Theory:Thm1}. 

\paragraph{Residual perturbation vs. DPSGD}
We have shown that the first residual perturbation~\eqref{eq:ResidualPerturbation1} outperforms the DPSGD in protecting membership privacy and improving classification accuracy. In this part, we further show that the second residual perturbation~\eqref{eq:ResidualPerturbation2} also outperforms the benchmark DPSGD with the above settings. Table~\ref{tab:RP:vs:DPSGD} lists the AUC of the attack model and training \& test accuracy of the target model; we see that the second residual perturbation can also improve the classification accuracy and {protecting comparable privacy}.

\begin{table}[t!]
\caption{\small Residual perturbation \eqref{eq:ResidualPerturbation2} {(\textbf{Strategy II})} vs. DPSGD in training ResNet8 for the IDC dataset classification. Ensemble of ResNet8 with residual perturbation is more accurate for classification (higher test acc) and protects comparable membership privacy compared to DPSGD. 
}\label{tab:RP:vs:DPSGD}
\centering
\fontsize{9.0pt}{0.9em}\selectfont
\begin{tabular}{c|c|c|c}
\hline
Model  & AUC & Training Acc & Test Acc\\
\hline
ResNet8 (DPSGD) & 0.503 & 0.831 & 0.828\\
En$_1$ResNet8   & 0.509 & 0.880 & 0.868\\
En$_5$ResNet8   & \bf 0.500 & \bf 0.895 & \bf 0.872\\
\hline
\end{tabular}
\end{table}

\subsubsection{Experiments on the CIFAR10/CIFAR100 Datasets}
\begin{figure}[!ht]
\centering
\begin{tabular}{cc}
\hskip -0.45cm\includegraphics[clip, trim=0cm 0cm 0cm 0.8cm,width=0.48\columnwidth]{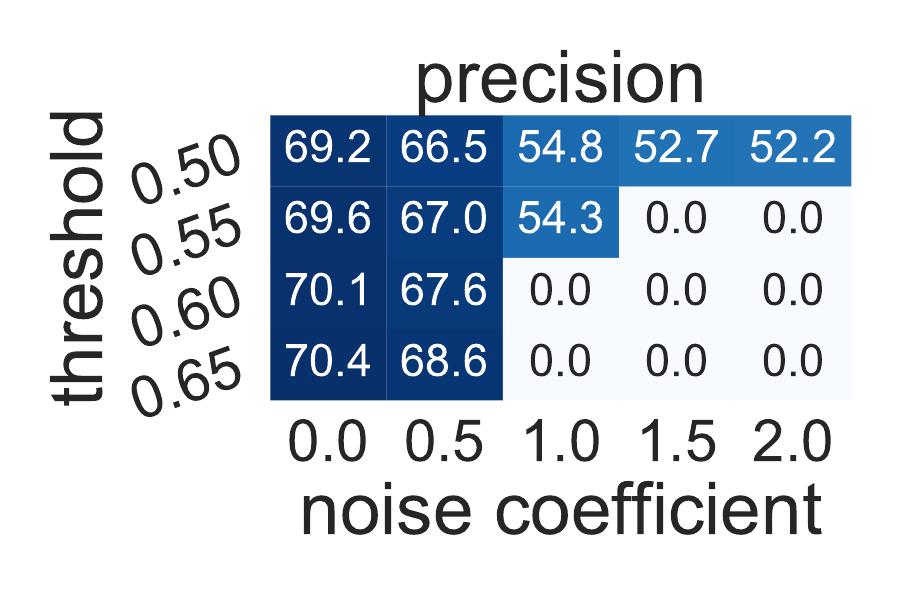}&
\hskip -0.45cm\includegraphics[clip, trim=0cm 0cm 0cm 0.8cm,width=0.48\columnwidth]{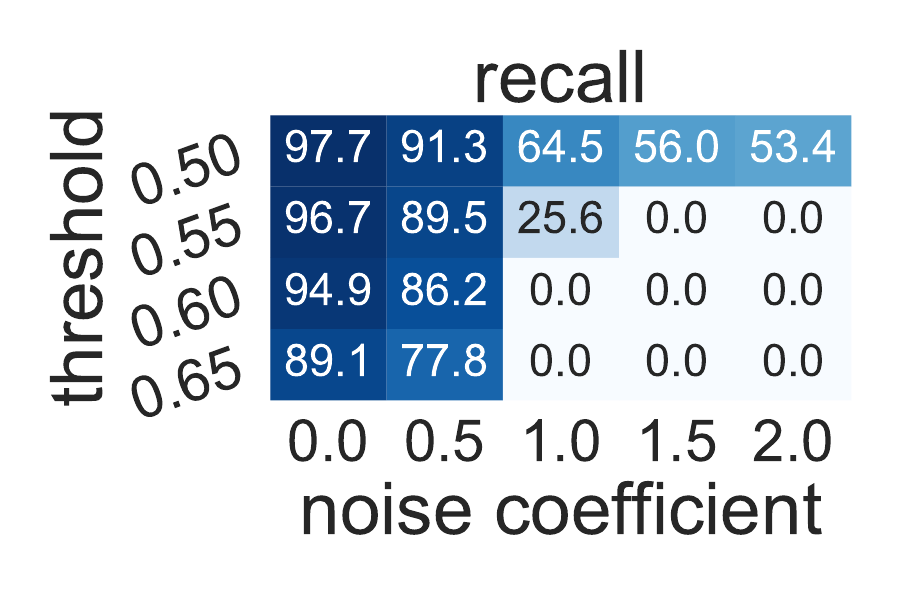}\\
\hskip -0.45cm\includegraphics[clip, trim=0cm 0cm 0cm 0.8cm,width=0.48\columnwidth]{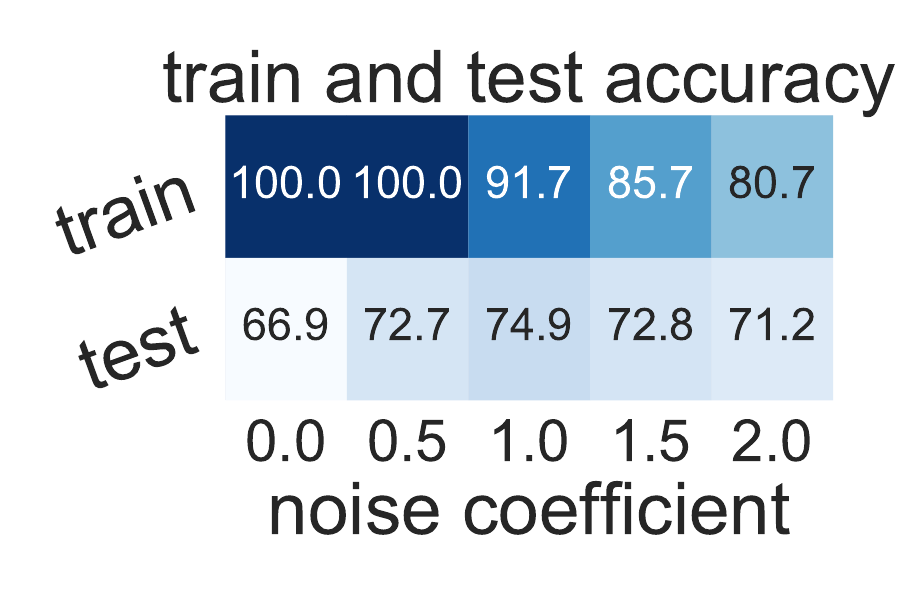}&
\hskip -0.45cm\includegraphics[clip, trim=0cm 0cm 0cm 0.8cm,width=0.48\columnwidth]{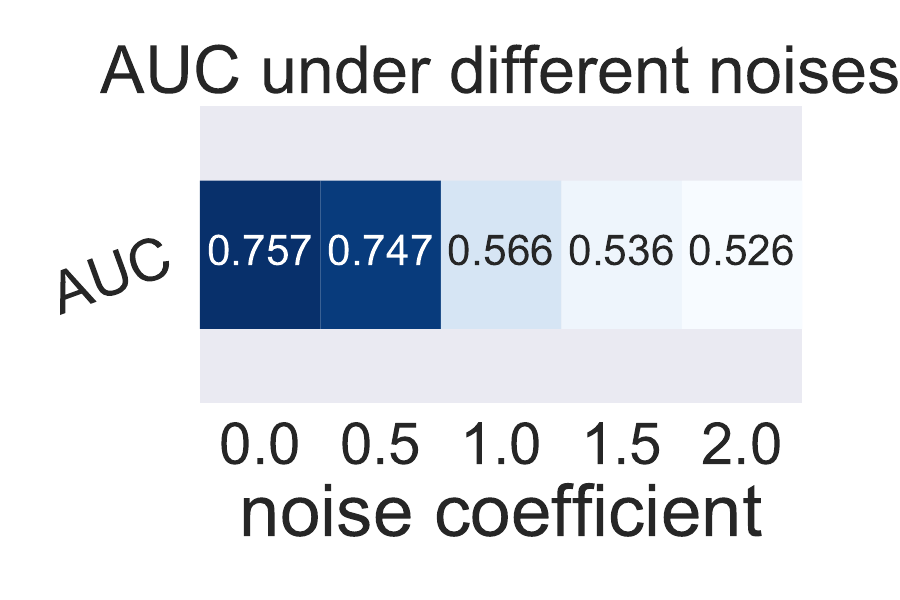}\\
\end{tabular}
\vskip -0.3cm
\caption{\small 
Performance of En$_5$ResNet8 with residual perturbation~\eqref{eq:ResidualPerturbation2} {(\textbf{Strategy II})} using different noise coefficients ($\gamma$) and membership inference attack threshold on CIFAR10. Residual perturbation~\eqref{eq:ResidualPerturbation2} can not only enhance the membership privacy, but also improve the classification accuracy. $\gamma=0$ corresponding to the baseline ResNet8 without residual perturbation or model ensemble.
(Unit: \%)}
\label{fig:Cifar10-WithSkip}\vspace{-0.4cm}
\end{figure}

\begin{figure}[!ht]
\centering
\begin{tabular}{cc}
\hskip -0.45cm\includegraphics[clip, trim=0cm 0cm 0cm 0cm,width=0.48\columnwidth]{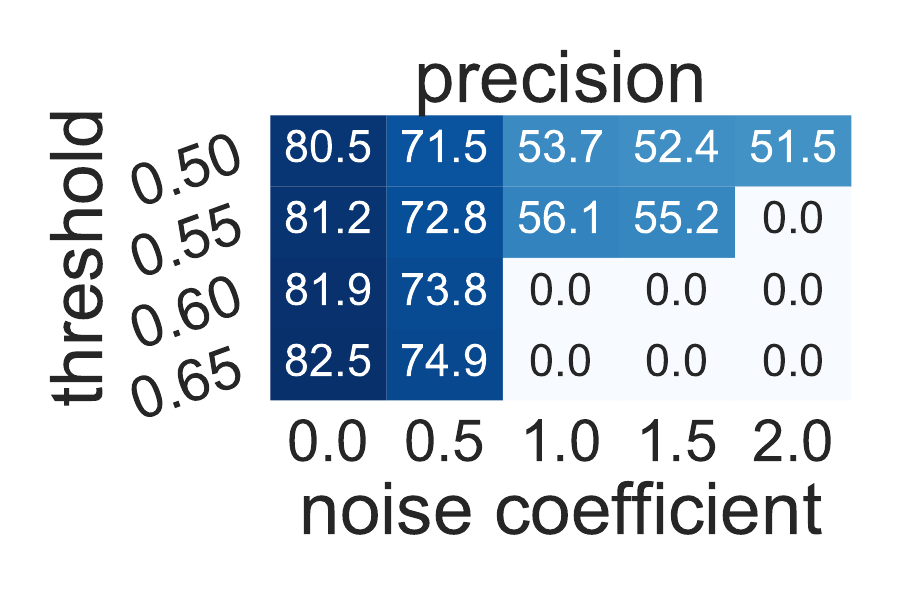}&
\hskip -0.45cm\includegraphics[clip, trim=0cm 0cm 0cm 0cm,width=0.48\columnwidth]{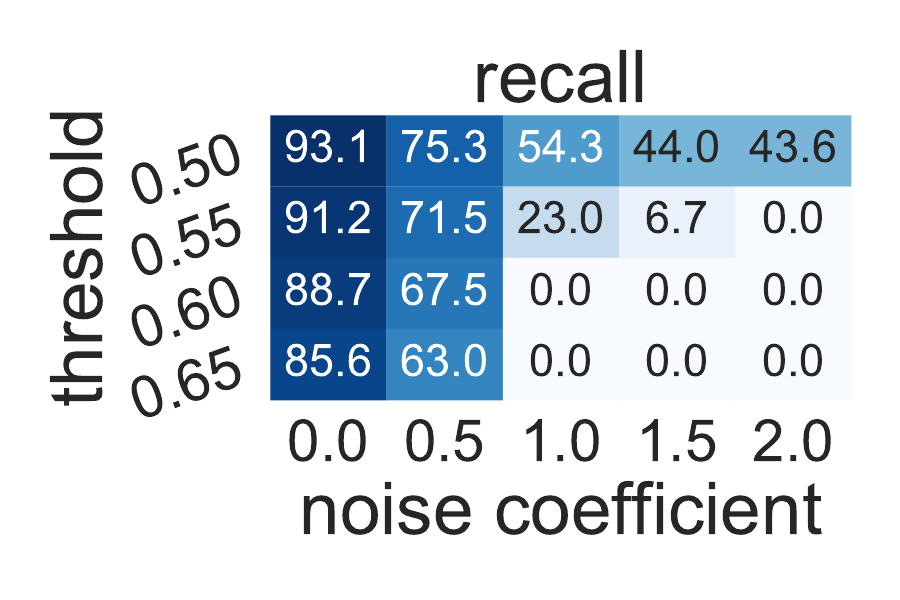}\\
\hskip -0.45cm\includegraphics[clip, trim=0cm 0cm 0cm 0cm,width=0.48\columnwidth]{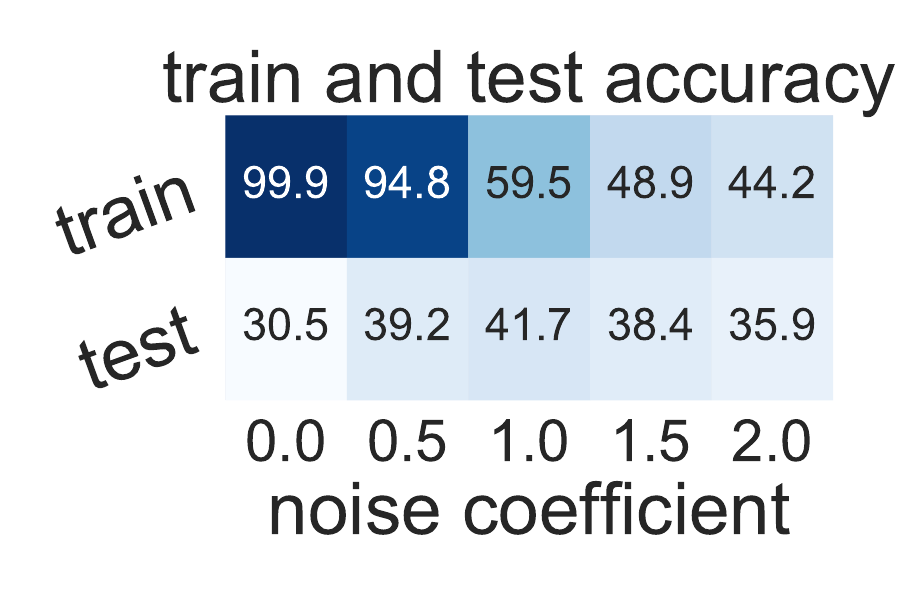}&
\hskip -0.45cm\includegraphics[clip, trim=0cm 0cm 0cm 0cm,width=0.48\columnwidth]{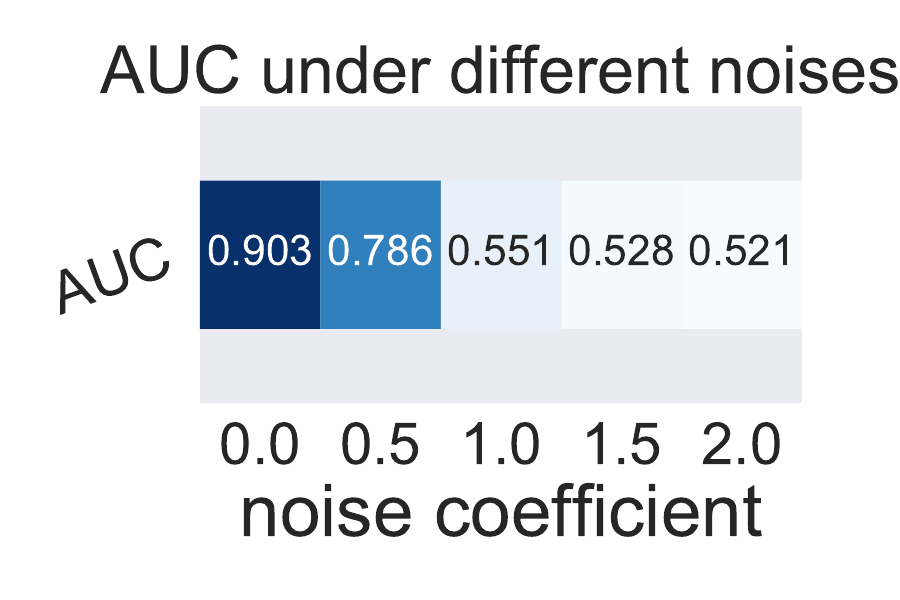}\\
\end{tabular}\vspace{-0.2cm}
\caption{\small 
Performance of En$_5$ResNet8 with residual perturbation~\eqref{eq:ResidualPerturbation2} {(\textbf{Strategy II})} using different noise coefficients ($\gamma$) and membership inference attack threshold on CIFAR100. Again, residual perturbation~\eqref{eq:ResidualPerturbation2} can not only enhance the membership privacy, but also improve the classification accuracy. $\gamma=0$ corresponding to the baseline ResNet8 without residual perturbation or model ensemble.
(Unit: \%)}
\label{fig:Cifar100}\vspace{-0.4cm}
\end{figure}

In this subsection, we will test the second residual perturbation~\eqref{eq:ResidualPerturbation2} on the CIFAR10/CIFAR100 datasets with the same model using the same settings as before. Fig.~\ref{fig:Cifar10-WithSkip} plots the performance of En$_5$ResNet8 on the CIFAR10 dataset. These results show that the ensemble of ResNets with residual perturbation~\eqref{eq:ResidualPerturbation2} is significantly more robust to the membership inference attack. For instance, the AUC of the attack model for ResNet8 and En$_5$ResNet8 ($\gamma=2.0$) is $0.757$ and $0.526$, respectively. Also, the classification accuracy of En$_5$ResNet8 ($\gamma=2.0$) is higher than that of ResNet8, and their accuracy is $71.2$\% and $66.9$\% for CIFAR10 classification. Fig.~\ref{fig:Cifar100} shows the results of En$_5$ResNet8 for CIFAR100 classification. These results confirm that residual perturbation~\eqref{eq:ResidualPerturbation2} can protect membership privacy and improve classification accuracy again. 

\subsubsection{ROC Curves for the Experiments on Different Datasets}\label{subsection:ROC_cn}
\begin{figure}[!ht]
\centering
\begin{tabular}{ccc}
\hskip -0.5cm\includegraphics[clip, trim=0cm 2cm 0cm 0.15cm,width=0.35\columnwidth]{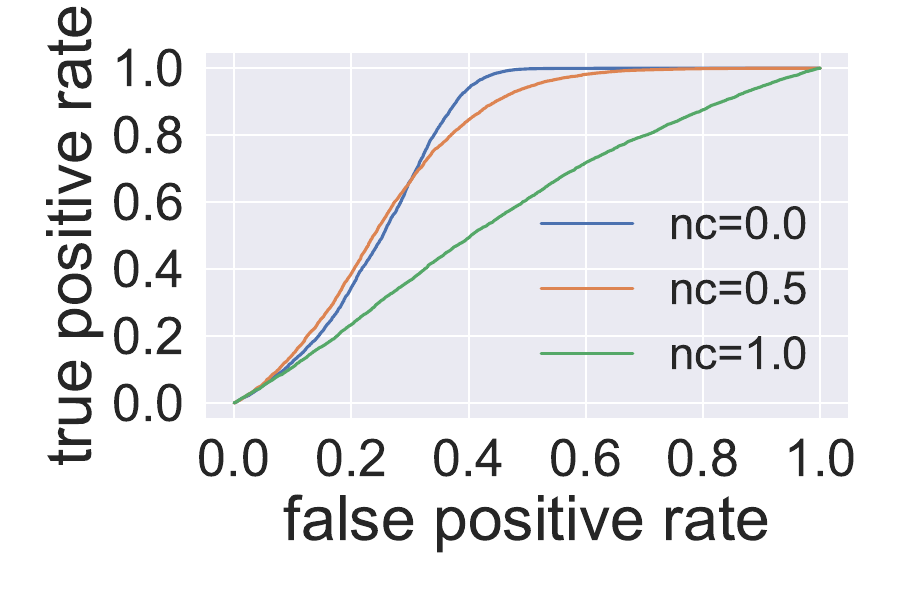}&
\hskip -0.5cm\includegraphics[clip, trim=0cm 2cm 0cm 0.15cm,width=0.35\columnwidth]{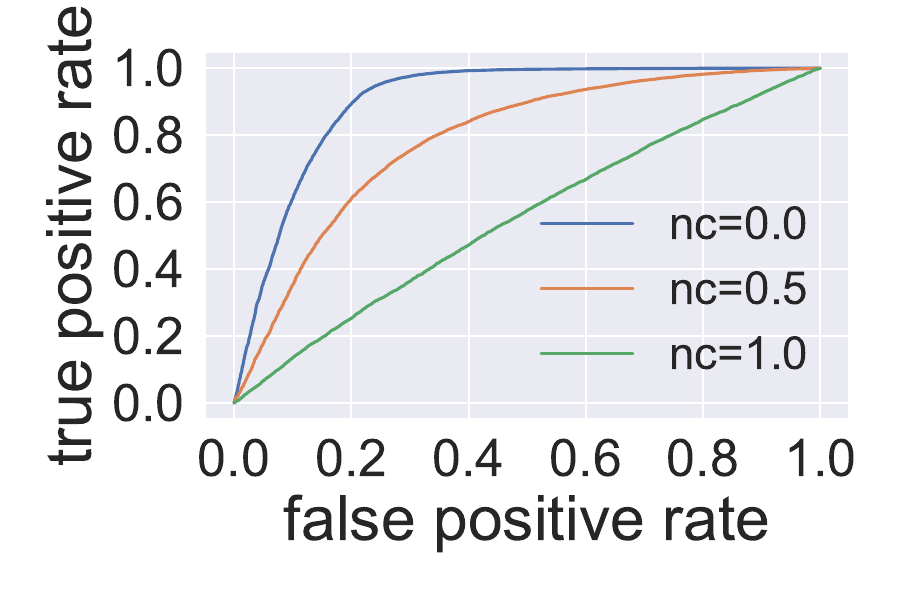}&
\hskip -0.5cm\includegraphics[clip, trim=0cm 2cm 0cm 0.15cm,width=0.35\columnwidth]{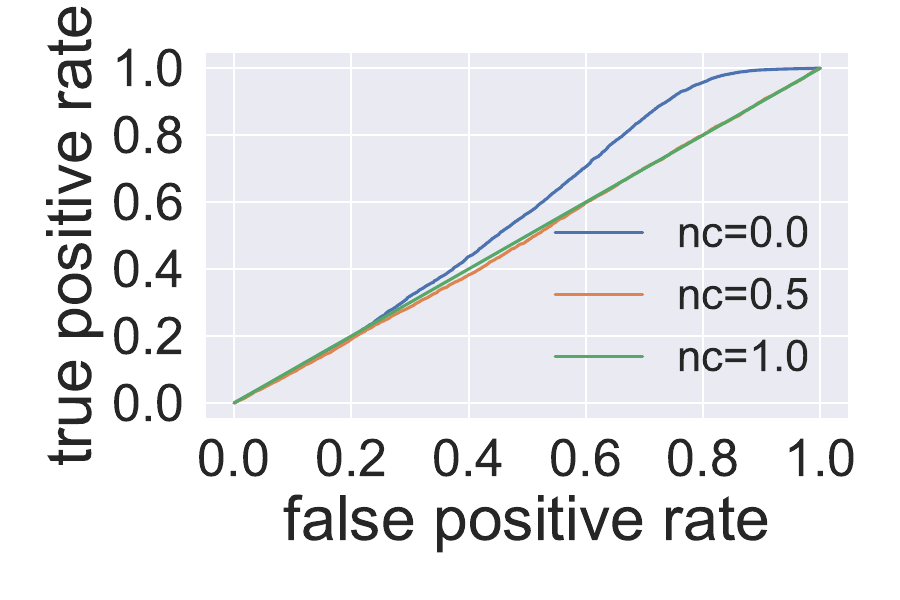}\\
{\small \hskip -0.2cm (a) CIFAR10} & {\small \hskip -0.2cm (b) CIFAR100} & {\small \hskip -0.2cm (c) IDC}\\
{\small \hskip -0.2cm (En$_5$ResNet8)} & {\small \hskip -0.2cm (En$_5$ResNet8)} & {\small \hskip -0.2cm (En$_5$ResNet8)}\\
\end{tabular}\vspace{-0.2cm}
\caption{\small ROC curves for residual perturbation~\eqref{eq:ResidualPerturbation2} {(\textbf{Strategy II})} for different datasets. (nc: noise coefficient)}
\label{fig:ROC}\vspace{-0.2cm}
\end{figure}

Fig.~\ref{fig:ROC} plots the ROC curves for the experiments on different datasets with different models using {\bf Strategy II}. 
These ROC curves again show that if $\gamma$ is sufficiently large, the attack model's prediction will be nearly a random guess.

\section{More Experimental Details}\label{sec:appendix:exp:details}
We give the detailed construction of the velocity field $F(\vx(t), \mW(t))$ in \eqref{Algorithm:Eq4} and \eqref{Algorithm:Eq3} that used to generate Figs.~\ref{fig:ODE:vs:SDE-Privacy} and \ref{fig:ODE:vs:SDE-Privacy-sde2} in Algorithm~\ref{alg:F}.
\begin{algorithm}[!ht]\small
\caption{The expression of $F(\vx(t), \mW(t))$}\label{alg:F}
\begin{algorithmic}
\State \textbf{Input: } image=$\vx(t)$; rows, cols, channels = image.shape.
\State \textbf{Output:} $F(\vx(t), \mW(t))$=noise injected image (NII).
\For {$k$ in range(channels)}
\For {$i$ in range(rows):}
\For {$j$ in range(cols):}
\State offset($j$) =int$( [j+50.0\text{cos}(2\pi i / 180)]\%$cols);
\State offset(i) =int$( [i+50.0\text{sin}(2\pi i / 180)]\%$row);
\State NII$[i, j, k] = \text{image}[(i + \text{offset}(j)) \% \text{rows},
(j + \text{offset}(i)) \% \text{cols}, k]$;
\EndFor 
\EndFor
\EndFor
\Return NII 
\end{algorithmic}
\end{algorithm}

\section{Concluding Remarks}\label{sec:Conclusion}
We proposed residual perturbations, whose theoretical foundation lies in the theory of stochastic differential equations, to protect data privacy for deep learning. Theoretically, we prove that the residual perturbation can reduce the generalization gap with differential privacy guarantees. Numerically, we have shown that residual perturbations are effective in protecting membership privacy on some benchmark datasets. {The proposed residual perturbation provides a feasible avenue for developing machine learning models that are more private and accurate than the baseline approach. Nevertheless, the current theoretical differential privacy budget is far from tight; developing analytical tools for analyzing a tighter privacy budget is an interesting future direction.}

\bibliographystyle{ieeetr}
\bibliography{references.bib}

\begin{IEEEbiographynophoto}{Wenqi Tao}
is a Ph.D. student in mathematics at Tsinghua University. His research focuses on graph-based machine learning and deep learning.
\end{IEEEbiographynophoto}\vspace{-0.5cm}

\begin{IEEEbiographynophoto}{Huaming Ling}
is a Ph.D. student in mathematics at Tsinghua University. His research focuses on deep learning and stochastic optimization algorithms.
\end{IEEEbiographynophoto}\vspace{-0.5cm}

\begin{IEEEbiographynophoto}{Zuoqiang Shi}
received Ph.D. in 2008 from Tsinghua University. He is an associate professor of mathematics at Tsinghua University.
\end{IEEEbiographynophoto}\vspace{-0.5cm}

\begin{IEEEbiographynophoto}{Bao Wang}
received Ph.D. in 2016 from Michigan State University. He is an assistant professor of mathematics at the University of Utah. He is a recipient of the Chancellor's award for postdoc research at UCLA. His research interests include scientific computing and deep learning.
\end{IEEEbiographynophoto}

\end{document}